\documentclass[letterpaper, 10 pt, conference]{ieeeconf} 
\IEEEoverridecommandlockouts  
\usepackage{pstricks}
\usepackage{ifplatform}
\usepackage{xkeyval}
\usepackage{rotating}

\usepackage{graphics} 
\usepackage{multirow}
\usepackage[off]{auto-pst-pdf}

\let\labelindent\relax

\usepackage{enumitem}
\setenumerate{topsep=-\baselineskip}
\usepackage{times} 
\usepackage{amsmath} 
\usepackage{psfrag}
\usepackage{cancel}

\usepackage[normalem]{ulem}
\usepackage{cite}
\usepackage{url}
\usepackage{color}
\usepackage{siunitx}

\definecolor{AliceBlue}{rgb}{0.94,0.97,1.00}
\definecolor{AntiqueWhite1}{rgb}{1.00,0.94,0.86}
\definecolor{AntiqueWhite2}{rgb}{0.93,0.87,0.80}
\definecolor{AntiqueWhite3}{rgb}{0.80,0.75,0.69}
\definecolor{AntiqueWhite4}{rgb}{0.55,0.51,0.47}
\definecolor{AntiqueWhite}{rgb}{0.98,0.92,0.84}
\definecolor{BlanchedAlmond}{rgb}{1.00,0.92,0.80}
\definecolor{BlueViolet}{rgb}{0.54,0.17,0.89}
\definecolor{CadetBlue1}{rgb}{0.60,0.96,1.00}
\definecolor{CadetBlue2}{rgb}{0.56,0.90,0.93}
\definecolor{CadetBlue3}{rgb}{0.48,0.77,0.80}
\definecolor{CadetBlue4}{rgb}{0.33,0.53,0.55}
\definecolor{CadetBlue}{rgb}{0.37,0.62,0.63}
\definecolor{CornflowerBlue}{rgb}{0.39,0.58,0.93}
\definecolor{DarkBlue}{rgb}{0.00,0.00,0.55}
\definecolor{DarkCyan}{rgb}{0.00,0.55,0.55}
\definecolor{DarkGoldenrod1}{rgb}{1.00,0.73,0.06}
\definecolor{DarkGoldenrod2}{rgb}{0.93,0.68,0.05}
\definecolor{DarkGoldenrod3}{rgb}{0.80,0.58,0.05}
\definecolor{DarkGoldenrod4}{rgb}{0.55,0.40,0.03}
\definecolor{DarkGoldenrod}{rgb}{0.72,0.53,0.04}
\definecolor{DarkGray}{rgb}{0.66,0.66,0.66}
\definecolor{DarkGreen}{rgb}{0.00,0.39,0.00}
\definecolor{DarkGrey}{rgb}{0.66,0.66,0.66}
\definecolor{DarkKhaki}{rgb}{0.74,0.72,0.42}
\definecolor{DarkMagenta}{rgb}{0.55,0.00,0.55}
\definecolor{DarkOliveGreen1}{rgb}{0.79,1.00,0.44}
\definecolor{DarkOliveGreen2}{rgb}{0.74,0.93,0.41}
\definecolor{DarkOliveGreen3}{rgb}{0.64,0.80,0.35}
\definecolor{DarkOliveGreen4}{rgb}{0.43,0.55,0.24}
\definecolor{DarkOliveGreen}{rgb}{0.33,0.42,0.18}
\definecolor{DarkOrange1}{rgb}{1.00,0.50,0.00}
\definecolor{DarkOrange2}{rgb}{0.93,0.46,0.00}
\definecolor{DarkOrange3}{rgb}{0.80,0.40,0.00}
\definecolor{DarkOrange4}{rgb}{0.55,0.27,0.00}
\definecolor{DarkOrange}{rgb}{1.00,0.55,0.00}
\definecolor{DarkOrchid1}{rgb}{0.75,0.24,1.00}
\definecolor{DarkOrchid2}{rgb}{0.70,0.23,0.93}
\definecolor{DarkOrchid3}{rgb}{0.60,0.20,0.80}
\definecolor{DarkOrchid4}{rgb}{0.41,0.13,0.55}
\definecolor{DarkOrchid}{rgb}{0.60,0.20,0.80}
\definecolor{DarkRed}{rgb}{0.55,0.00,0.00}
\definecolor{DarkSalmon}{rgb}{0.91,0.59,0.48}
\definecolor{DarkSeaGreen1}{rgb}{0.76,1.00,0.76}
\definecolor{DarkSeaGreen2}{rgb}{0.71,0.93,0.71}
\definecolor{DarkSeaGreen3}{rgb}{0.61,0.80,0.61}
\definecolor{DarkSeaGreen4}{rgb}{0.41,0.55,0.41}
\definecolor{DarkSeaGreen}{rgb}{0.56,0.74,0.56}
\definecolor{DarkSlateBlue}{rgb}{0.28,0.24,0.55}
\definecolor{DarkSlateGray1}{rgb}{0.59,1.00,1.00}
\definecolor{DarkSlateGray2}{rgb}{0.55,0.93,0.93}
\definecolor{DarkSlateGray3}{rgb}{0.47,0.80,0.80}
\definecolor{DarkSlateGray4}{rgb}{0.32,0.55,0.55}
\definecolor{DarkSlateGray}{rgb}{0.18,0.31,0.31}
\definecolor{DarkSlateGrey}{rgb}{0.18,0.31,0.31}
\definecolor{DarkTurquoise}{rgb}{0.00,0.81,0.82}
\definecolor{DarkViolet}{rgb}{0.58,0.00,0.83}
\definecolor{DeepPink1}{rgb}{1.00,0.08,0.58}
\definecolor{DeepPink2}{rgb}{0.93,0.07,0.54}
\definecolor{DeepPink3}{rgb}{0.80,0.06,0.46}
\definecolor{DeepPink4}{rgb}{0.55,0.04,0.31}
\definecolor{DeepPink}{rgb}{1.00,0.08,0.58}
\definecolor{DeepSkyBlue1}{rgb}{0.00,0.75,1.00}
\definecolor{DeepSkyBlue2}{rgb}{0.00,0.70,0.93}
\definecolor{DeepSkyBlue3}{rgb}{0.00,0.60,0.80}
\definecolor{DeepSkyBlue4}{rgb}{0.00,0.41,0.55}
\definecolor{DeepSkyBlue}{rgb}{0.00,0.75,1.00}
\definecolor{DimGray}{rgb}{0.41,0.41,0.41}
\definecolor{DimGrey}{rgb}{0.41,0.41,0.41}
\definecolor{DodgerBlue1}{rgb}{0.12,0.56,1.00}
\definecolor{DodgerBlue2}{rgb}{0.11,0.53,0.93}
\definecolor{DodgerBlue3}{rgb}{0.09,0.45,0.80}
\definecolor{DodgerBlue4}{rgb}{0.06,0.31,0.55}
\definecolor{DodgerBlue}{rgb}{0.12,0.56,1.00}
\definecolor{FloralWhite}{rgb}{1.00,0.98,0.94}
\definecolor{ForestGreen}{rgb}{0.13,0.55,0.13}
\definecolor{GhostWhite}{rgb}{0.97,0.97,1.00}
\definecolor{GreenYellow}{rgb}{0.68,1.00,0.18}
\definecolor{HotPink1}{rgb}{1.00,0.43,0.71}
\definecolor{HotPink2}{rgb}{0.93,0.42,0.65}
\definecolor{HotPink3}{rgb}{0.80,0.38,0.56}
\definecolor{HotPink4}{rgb}{0.55,0.23,0.38}
\definecolor{HotPink}{rgb}{1.00,0.41,0.71}
\definecolor{IndianRed1}{rgb}{1.00,0.42,0.42}
\definecolor{IndianRed2}{rgb}{0.93,0.39,0.39}
\definecolor{IndianRed3}{rgb}{0.80,0.33,0.33}
\definecolor{IndianRed4}{rgb}{0.55,0.23,0.23}
\definecolor{IndianRed}{rgb}{0.80,0.36,0.36}
\definecolor{LavenderBlush1}{rgb}{1.00,0.94,0.96}
\definecolor{LavenderBlush2}{rgb}{0.93,0.88,0.90}
\definecolor{LavenderBlush3}{rgb}{0.80,0.76,0.77}
\definecolor{LavenderBlush4}{rgb}{0.55,0.51,0.53}
\definecolor{LavenderBlush}{rgb}{1.00,0.94,0.96}
\definecolor{LawnGreen}{rgb}{0.49,0.99,0.00}
\definecolor{LemonChiffon1}{rgb}{1.00,0.98,0.80}
\definecolor{LemonChiffon2}{rgb}{0.93,0.91,0.75}
\definecolor{LemonChiffon3}{rgb}{0.80,0.79,0.65}
\definecolor{LemonChiffon4}{rgb}{0.55,0.54,0.44}
\definecolor{LemonChiffon}{rgb}{1.00,0.98,0.80}
\definecolor{LightBlue1}{rgb}{0.75,0.94,1.00}
\definecolor{LightBlue2}{rgb}{0.70,0.87,0.93}
\definecolor{LightBlue3}{rgb}{0.60,0.75,0.80}
\definecolor{LightBlue4}{rgb}{0.41,0.51,0.55}
\definecolor{LightBlue}{rgb}{0.68,0.85,0.90}
\definecolor{LightCoral}{rgb}{0.94,0.50,0.50}
\definecolor{LightCyan1}{rgb}{0.88,1.00,1.00}
\definecolor{LightCyan2}{rgb}{0.82,0.93,0.93}
\definecolor{LightCyan3}{rgb}{0.71,0.80,0.80}
\definecolor{LightCyan4}{rgb}{0.48,0.55,0.55}
\definecolor{LightCyan}{rgb}{0.88,1.00,1.00}
\definecolor{LightGoldenrod1}{rgb}{1.00,0.93,0.55}
\definecolor{LightGoldenrod2}{rgb}{0.93,0.86,0.51}
\definecolor{LightGoldenrod3}{rgb}{0.80,0.75,0.44}
\definecolor{LightGoldenrod4}{rgb}{0.55,0.51,0.30}
\definecolor{LightGoldenrodYellow}{rgb}{0.98,0.98,0.82}
\definecolor{LightGoldenrod}{rgb}{0.93,0.87,0.51}
\definecolor{LightGray}{rgb}{0.83,0.83,0.83}
\definecolor{LightGreen}{rgb}{0.56,0.93,0.56}
\definecolor{LightGrey}{rgb}{0.83,0.83,0.83}
\definecolor{LightPink1}{rgb}{1.00,0.68,0.73}
\definecolor{LightPink2}{rgb}{0.93,0.64,0.68}
\definecolor{LightPink3}{rgb}{0.80,0.55,0.58}
\definecolor{LightPink4}{rgb}{0.55,0.37,0.40}
\definecolor{LightPink}{rgb}{1.00,0.71,0.76}
\definecolor{LightSalmon1}{rgb}{1.00,0.63,0.48}
\definecolor{LightSalmon2}{rgb}{0.93,0.58,0.45}
\definecolor{LightSalmon3}{rgb}{0.80,0.51,0.38}
\definecolor{LightSalmon4}{rgb}{0.55,0.34,0.26}
\definecolor{LightSalmon}{rgb}{1.00,0.63,0.48}
\definecolor{LightSeaGreen}{rgb}{0.13,0.70,0.67}
\definecolor{LightSkyBlue1}{rgb}{0.69,0.89,1.00}
\definecolor{LightSkyBlue2}{rgb}{0.64,0.83,0.93}
\definecolor{LightSkyBlue3}{rgb}{0.55,0.71,0.80}
\definecolor{LightSkyBlue4}{rgb}{0.38,0.48,0.55}
\definecolor{LightSkyBlue}{rgb}{0.53,0.81,0.98}
\definecolor{LightSlateBlue}{rgb}{0.52,0.44,1.00}
\definecolor{LightSlateGray}{rgb}{0.47,0.53,0.60}
\definecolor{LightSlateGrey}{rgb}{0.47,0.53,0.60}
\definecolor{LightSteelBlue1}{rgb}{0.79,0.88,1.00}
\definecolor{LightSteelBlue2}{rgb}{0.74,0.82,0.93}
\definecolor{LightSteelBlue3}{rgb}{0.64,0.71,0.80}
\definecolor{LightSteelBlue4}{rgb}{0.43,0.48,0.55}
\definecolor{LightSteelBlue}{rgb}{0.69,0.77,0.87}
\definecolor{LightYellow1}{rgb}{1.00,1.00,0.88}
\definecolor{LightYellow2}{rgb}{0.93,0.93,0.82}
\definecolor{LightYellow3}{rgb}{0.80,0.80,0.71}
\definecolor{LightYellow4}{rgb}{0.55,0.55,0.48}
\definecolor{LightYellow}{rgb}{1.00,1.00,0.88}
\definecolor{LimeGreen}{rgb}{0.20,0.80,0.20}
\definecolor{MediumAquamarine}{rgb}{0.40,0.80,0.67}
\definecolor{MediumBlue}{rgb}{0.00,0.00,0.80}
\definecolor{MediumOrchid1}{rgb}{0.88,0.40,1.00}
\definecolor{MediumOrchid2}{rgb}{0.82,0.37,0.93}
\definecolor{MediumOrchid3}{rgb}{0.71,0.32,0.80}
\definecolor{MediumOrchid4}{rgb}{0.48,0.22,0.55}
\definecolor{MediumOrchid}{rgb}{0.73,0.33,0.83}
\definecolor{MediumPurple1}{rgb}{0.67,0.51,1.00}
\definecolor{MediumPurple2}{rgb}{0.62,0.47,0.93}
\definecolor{MediumPurple3}{rgb}{0.54,0.41,0.80}
\definecolor{MediumPurple4}{rgb}{0.36,0.28,0.55}
\definecolor{MediumPurple}{rgb}{0.58,0.44,0.86}
\definecolor{MediumSeaGreen}{rgb}{0.24,0.70,0.44}
\definecolor{MediumSlateBlue}{rgb}{0.48,0.41,0.93}
\definecolor{MediumSpringGreen}{rgb}{0.00,0.98,0.60}
\definecolor{MediumTurquoise}{rgb}{0.28,0.82,0.80}
\definecolor{MediumVioletRed}{rgb}{0.78,0.08,0.52}
\definecolor{MidnightBlue}{rgb}{0.10,0.10,0.44}
\definecolor{MintCream}{rgb}{0.96,1.00,0.98}
\definecolor{MistyRose1}{rgb}{1.00,0.89,0.88}
\definecolor{MistyRose2}{rgb}{0.93,0.84,0.82}
\definecolor{MistyRose3}{rgb}{0.80,0.72,0.71}
\definecolor{MistyRose4}{rgb}{0.55,0.49,0.48}
\definecolor{MistyRose}{rgb}{1.00,0.89,0.88}
\definecolor{NavajoWhite1}{rgb}{1.00,0.87,0.68}
\definecolor{NavajoWhite2}{rgb}{0.93,0.81,0.63}
\definecolor{NavajoWhite3}{rgb}{0.80,0.70,0.55}
\definecolor{NavajoWhite4}{rgb}{0.55,0.47,0.37}
\definecolor{NavajoWhite}{rgb}{1.00,0.87,0.68}
\definecolor{NavyBlue}{rgb}{0.00,0.00,0.50}
\definecolor{OldLace}{rgb}{0.99,0.96,0.90}
\definecolor{OliveDrab1}{rgb}{0.75,1.00,0.24}
\definecolor{OliveDrab2}{rgb}{0.70,0.93,0.23}
\definecolor{OliveDrab3}{rgb}{0.60,0.80,0.20}
\definecolor{OliveDrab4}{rgb}{0.41,0.55,0.13}
\definecolor{OliveDrab}{rgb}{0.42,0.56,0.14}
\definecolor{OrangeRed1}{rgb}{1.00,0.27,0.00}
\definecolor{OrangeRed2}{rgb}{0.93,0.25,0.00}
\definecolor{OrangeRed3}{rgb}{0.80,0.22,0.00}
\definecolor{OrangeRed4}{rgb}{0.55,0.15,0.00}
\definecolor{OrangeRed}{rgb}{1.00,0.27,0.00}
\definecolor{PaleGoldenrod}{rgb}{0.93,0.91,0.67}
\definecolor{PaleGreen1}{rgb}{0.60,1.00,0.60}
\definecolor{PaleGreen2}{rgb}{0.56,0.93,0.56}
\definecolor{PaleGreen3}{rgb}{0.49,0.80,0.49}
\definecolor{PaleGreen4}{rgb}{0.33,0.55,0.33}
\definecolor{PaleGreen}{rgb}{0.60,0.98,0.60}
\definecolor{PaleTurquoise1}{rgb}{0.73,1.00,1.00}
\definecolor{PaleTurquoise2}{rgb}{0.68,0.93,0.93}
\definecolor{PaleTurquoise3}{rgb}{0.59,0.80,0.80}
\definecolor{PaleTurquoise4}{rgb}{0.40,0.55,0.55}
\definecolor{PaleTurquoise}{rgb}{0.69,0.93,0.93}
\definecolor{PaleVioletRed1}{rgb}{1.00,0.51,0.67}
\definecolor{PaleVioletRed2}{rgb}{0.93,0.47,0.62}
\definecolor{PaleVioletRed3}{rgb}{0.80,0.41,0.54}
\definecolor{PaleVioletRed4}{rgb}{0.55,0.28,0.36}
\definecolor{PaleVioletRed}{rgb}{0.86,0.44,0.58}
\definecolor{PapayaWhip}{rgb}{1.00,0.94,0.84}
\definecolor{PeachPuff1}{rgb}{1.00,0.85,0.73}
\definecolor{PeachPuff2}{rgb}{0.93,0.80,0.68}
\definecolor{PeachPuff3}{rgb}{0.80,0.69,0.58}
\definecolor{PeachPuff4}{rgb}{0.55,0.47,0.40}
\definecolor{PeachPuff}{rgb}{1.00,0.85,0.73}
\definecolor{PowderBlue}{rgb}{0.69,0.88,0.90}
\definecolor{RosyBrown1}{rgb}{1.00,0.76,0.76}
\definecolor{RosyBrown2}{rgb}{0.93,0.71,0.71}
\definecolor{RosyBrown3}{rgb}{0.80,0.61,0.61}
\definecolor{RosyBrown4}{rgb}{0.55,0.41,0.41}
\definecolor{RosyBrown}{rgb}{0.74,0.56,0.56}
\definecolor{RoyalBlue1}{rgb}{0.28,0.46,1.00}
\definecolor{RoyalBlue2}{rgb}{0.26,0.43,0.93}
\definecolor{RoyalBlue3}{rgb}{0.23,0.37,0.80}
\definecolor{RoyalBlue4}{rgb}{0.15,0.25,0.55}
\definecolor{RoyalBlue}{rgb}{0.25,0.41,0.88}
\definecolor{SaddleBrown}{rgb}{0.55,0.27,0.07}
\definecolor{SandyBrown}{rgb}{0.96,0.64,0.38}
\definecolor{SeaGreen1}{rgb}{0.33,1.00,0.62}
\definecolor{SeaGreen2}{rgb}{0.31,0.93,0.58}
\definecolor{SeaGreen3}{rgb}{0.26,0.80,0.50}
\definecolor{SeaGreen4}{rgb}{0.18,0.55,0.34}
\definecolor{SeaGreen}{rgb}{0.18,0.55,0.34}
\definecolor{SkyBlue1}{rgb}{0.53,0.81,1.00}
\definecolor{SkyBlue2}{rgb}{0.49,0.75,0.93}
\definecolor{SkyBlue3}{rgb}{0.42,0.65,0.80}
\definecolor{SkyBlue4}{rgb}{0.29,0.44,0.55}
\definecolor{SkyBlue}{rgb}{0.53,0.81,0.92}
\definecolor{SlateBlue1}{rgb}{0.51,0.44,1.00}
\definecolor{SlateBlue2}{rgb}{0.48,0.40,0.93}
\definecolor{SlateBlue3}{rgb}{0.41,0.35,0.80}
\definecolor{SlateBlue4}{rgb}{0.28,0.24,0.55}
\definecolor{SlateBlue}{rgb}{0.42,0.35,0.80}
\definecolor{SlateGray1}{rgb}{0.78,0.89,1.00}
\definecolor{SlateGray2}{rgb}{0.73,0.83,0.93}
\definecolor{SlateGray3}{rgb}{0.62,0.71,0.80}
\definecolor{SlateGray4}{rgb}{0.42,0.48,0.55}
\definecolor{SlateGray}{rgb}{0.44,0.50,0.56}
\definecolor{SlateGrey}{rgb}{0.44,0.50,0.56}
\definecolor{SpringGreen1}{rgb}{0.00,1.00,0.50}
\definecolor{SpringGreen2}{rgb}{0.00,0.93,0.46}
\definecolor{SpringGreen3}{rgb}{0.00,0.80,0.40}
\definecolor{SpringGreen4}{rgb}{0.00,0.55,0.27}
\definecolor{SpringGreen}{rgb}{0.00,1.00,0.50}
\definecolor{SteelBlue1}{rgb}{0.39,0.72,1.00}
\definecolor{SteelBlue2}{rgb}{0.36,0.67,0.93}
\definecolor{SteelBlue3}{rgb}{0.31,0.58,0.80}
\definecolor{SteelBlue4}{rgb}{0.21,0.39,0.55}
\definecolor{SteelBlue}{rgb}{0.27,0.51,0.71}
\definecolor{VioletRed1}{rgb}{1.00,0.24,0.59}
\definecolor{VioletRed2}{rgb}{0.93,0.23,0.55}
\definecolor{VioletRed3}{rgb}{0.80,0.20,0.47}
\definecolor{VioletRed4}{rgb}{0.55,0.13,0.32}
\definecolor{VioletRed}{rgb}{0.82,0.13,0.56}
\definecolor{WhiteSmoke}{rgb}{0.96,0.96,0.96}
\definecolor{YellowGreen}{rgb}{0.60,0.80,0.20}
\definecolor{aliceblue}{rgb}{0.94,0.97,1.00}
\definecolor{antiquewhite}{rgb}{0.98,0.92,0.84}
\definecolor{aquamarine1}{rgb}{0.50,1.00,0.83}
\definecolor{aquamarine2}{rgb}{0.46,0.93,0.78}
\definecolor{aquamarine3}{rgb}{0.40,0.80,0.67}
\definecolor{aquamarine4}{rgb}{0.27,0.55,0.45}
\definecolor{aquamarine}{rgb}{0.50,1.00,0.83}
\definecolor{azure1}{rgb}{0.94,1.00,1.00}
\definecolor{azure2}{rgb}{0.88,0.93,0.93}
\definecolor{azure3}{rgb}{0.76,0.80,0.80}
\definecolor{azure4}{rgb}{0.51,0.55,0.55}
\definecolor{azure}{rgb}{0.94,1.00,1.00}
\definecolor{beige}{rgb}{0.96,0.96,0.86}
\definecolor{bisque1}{rgb}{1.00,0.89,0.77}
\definecolor{bisque2}{rgb}{0.93,0.84,0.72}
\definecolor{bisque3}{rgb}{0.80,0.72,0.62}
\definecolor{bisque4}{rgb}{0.55,0.49,0.42}
\definecolor{bisque}{rgb}{1.00,0.89,0.77}
\definecolor{black}{rgb}{0.00,0.00,0.00}
\definecolor{blanchedalmond}{rgb}{1.00,0.92,0.80}
\definecolor{blue1}{rgb}{0.00,0.00,1.00}
\definecolor{blue2}{rgb}{0.00,0.00,0.93}
\definecolor{blue3}{rgb}{0.00,0.00,0.80}
\definecolor{blue4}{rgb}{0.00,0.00,0.55}
\definecolor{blueviolet}{rgb}{0.54,0.17,0.89}
\definecolor{blue}{rgb}{0.00,0.00,1.00}
\definecolor{brown1}{rgb}{1.00,0.25,0.25}
\definecolor{brown2}{rgb}{0.93,0.23,0.23}
\definecolor{brown3}{rgb}{0.80,0.20,0.20}
\definecolor{brown4}{rgb}{0.55,0.14,0.14}
\definecolor{brown}{rgb}{0.65,0.16,0.16}
\definecolor{burlywood1}{rgb}{1.00,0.83,0.61}
\definecolor{burlywood2}{rgb}{0.93,0.77,0.57}
\definecolor{burlywood3}{rgb}{0.80,0.67,0.49}
\definecolor{burlywood4}{rgb}{0.55,0.45,0.33}
\definecolor{burlywood}{rgb}{0.87,0.72,0.53}
\definecolor{cadetblue}{rgb}{0.37,0.62,0.63}
\definecolor{chartreuse1}{rgb}{0.50,1.00,0.00}
\definecolor{chartreuse2}{rgb}{0.46,0.93,0.00}
\definecolor{chartreuse3}{rgb}{0.40,0.80,0.00}
\definecolor{chartreuse4}{rgb}{0.27,0.55,0.00}
\definecolor{chartreuse}{rgb}{0.50,1.00,0.00}
\definecolor{chocolate1}{rgb}{1.00,0.50,0.14}
\definecolor{chocolate2}{rgb}{0.93,0.46,0.13}
\definecolor{chocolate3}{rgb}{0.80,0.40,0.11}
\definecolor{chocolate4}{rgb}{0.55,0.27,0.07}
\definecolor{chocolate}{rgb}{0.82,0.41,0.12}
\definecolor{coral1}{rgb}{1.00,0.45,0.34}
\definecolor{coral2}{rgb}{0.93,0.42,0.31}
\definecolor{coral3}{rgb}{0.80,0.36,0.27}
\definecolor{coral4}{rgb}{0.55,0.24,0.18}
\definecolor{coral}{rgb}{1.00,0.50,0.31}
\definecolor{cornflowerblue}{rgb}{0.39,0.58,0.93}
\definecolor{cornsilk1}{rgb}{1.00,0.97,0.86}
\definecolor{cornsilk2}{rgb}{0.93,0.91,0.80}
\definecolor{cornsilk3}{rgb}{0.80,0.78,0.69}
\definecolor{cornsilk4}{rgb}{0.55,0.53,0.47}
\definecolor{cornsilk}{rgb}{1.00,0.97,0.86}
\definecolor{cyan1}{rgb}{0.00,1.00,1.00}
\definecolor{cyan2}{rgb}{0.00,0.93,0.93}
\definecolor{cyan3}{rgb}{0.00,0.80,0.80}
\definecolor{cyan4}{rgb}{0.00,0.55,0.55}
\definecolor{cyan}{rgb}{0.00,1.00,1.00}
\definecolor{darkblue}{rgb}{0.00,0.00,0.55}
\definecolor{darkcyan}{rgb}{0.00,0.55,0.55}
\definecolor{darkgoldenrod}{rgb}{0.72,0.53,0.04}
\definecolor{darkgray}{rgb}{0.66,0.66,0.66}
\definecolor{darkgreen}{rgb}{0.00,0.39,0.00}
\definecolor{darkgrey}{rgb}{0.66,0.66,0.66}
\definecolor{darkkhaki}{rgb}{0.74,0.72,0.42}
\definecolor{darkmagenta}{rgb}{0.55,0.00,0.55}
\definecolor{darkolive}{rgb}{0.33,0.42,0.18}
\definecolor{darkorange}{rgb}{1.00,0.55,0.00}
\definecolor{darkorchid}{rgb}{0.60,0.20,0.80}
\definecolor{darkred}{rgb}{0.55,0.00,0.00}
\definecolor{darksalmon}{rgb}{0.91,0.59,0.48}
\definecolor{darksea}{rgb}{0.56,0.74,0.56}
\definecolor{darkslate}{rgb}{0.18,0.31,0.31}
\definecolor{darkslate}{rgb}{0.18,0.31,0.31}
\definecolor{darkslate}{rgb}{0.28,0.24,0.55}
\definecolor{darkturquoise}{rgb}{0.00,0.81,0.82}
\definecolor{darkviolet}{rgb}{0.58,0.00,0.83}
\definecolor{deeppink}{rgb}{1.00,0.08,0.58}
\definecolor{deepsky}{rgb}{0.00,0.75,1.00}
\definecolor{dimgray}{rgb}{0.41,0.41,0.41}
\definecolor{dimgrey}{rgb}{0.41,0.41,0.41}
\definecolor{dodgerblue}{rgb}{0.12,0.56,1.00}
\definecolor{firebrick1}{rgb}{1.00,0.19,0.19}
\definecolor{firebrick2}{rgb}{0.93,0.17,0.17}
\definecolor{firebrick3}{rgb}{0.80,0.15,0.15}
\definecolor{firebrick4}{rgb}{0.55,0.10,0.10}
\definecolor{firebrick}{rgb}{0.70,0.13,0.13}
\definecolor{floralwhite}{rgb}{1.00,0.98,0.94}
\definecolor{forestgreen}{rgb}{0.13,0.55,0.13}
\definecolor{gainsboro}{rgb}{0.86,0.86,0.86}
\definecolor{ghostwhite}{rgb}{0.97,0.97,1.00}
\definecolor{gold1}{rgb}{1.00,0.84,0.00}
\definecolor{gold2}{rgb}{0.93,0.79,0.00}
\definecolor{gold3}{rgb}{0.80,0.68,0.00}
\definecolor{gold4}{rgb}{0.55,0.46,0.00}
\definecolor{goldenrod1}{rgb}{1.00,0.76,0.15}
\definecolor{goldenrod2}{rgb}{0.93,0.71,0.13}
\definecolor{goldenrod3}{rgb}{0.80,0.61,0.11}
\definecolor{goldenrod4}{rgb}{0.55,0.41,0.08}
\definecolor{goldenrod}{rgb}{0.85,0.65,0.13}
\definecolor{gold}{rgb}{1.00,0.84,0.00}
\definecolor{gray0}{rgb}{0.00,0.00,0.00}
\definecolor{gray100}{rgb}{1.00,1.00,1.00}
\definecolor{gray10}{rgb}{0.10,0.10,0.10}
\definecolor{gray11}{rgb}{0.11,0.11,0.11}
\definecolor{gray12}{rgb}{0.12,0.12,0.12}
\definecolor{gray13}{rgb}{0.13,0.13,0.13}
\definecolor{gray14}{rgb}{0.14,0.14,0.14}
\definecolor{gray15}{rgb}{0.15,0.15,0.15}
\definecolor{gray16}{rgb}{0.16,0.16,0.16}
\definecolor{gray17}{rgb}{0.17,0.17,0.17}
\definecolor{gray18}{rgb}{0.18,0.18,0.18}
\definecolor{gray19}{rgb}{0.19,0.19,0.19}
\definecolor{gray1}{rgb}{0.01,0.01,0.01}
\definecolor{gray20}{rgb}{0.20,0.20,0.20}
\definecolor{gray21}{rgb}{0.21,0.21,0.21}
\definecolor{gray22}{rgb}{0.22,0.22,0.22}
\definecolor{gray23}{rgb}{0.23,0.23,0.23}
\definecolor{gray24}{rgb}{0.24,0.24,0.24}
\definecolor{gray25}{rgb}{0.25,0.25,0.25}
\definecolor{gray26}{rgb}{0.26,0.26,0.26}
\definecolor{gray27}{rgb}{0.27,0.27,0.27}
\definecolor{gray28}{rgb}{0.28,0.28,0.28}
\definecolor{gray29}{rgb}{0.29,0.29,0.29}
\definecolor{gray2}{rgb}{0.02,0.02,0.02}
\definecolor{gray30}{rgb}{0.30,0.30,0.30}
\definecolor{gray31}{rgb}{0.31,0.31,0.31}
\definecolor{gray32}{rgb}{0.32,0.32,0.32}
\definecolor{gray33}{rgb}{0.33,0.33,0.33}
\definecolor{gray34}{rgb}{0.34,0.34,0.34}
\definecolor{gray35}{rgb}{0.35,0.35,0.35}
\definecolor{gray36}{rgb}{0.36,0.36,0.36}
\definecolor{gray37}{rgb}{0.37,0.37,0.37}
\definecolor{gray38}{rgb}{0.38,0.38,0.38}
\definecolor{gray39}{rgb}{0.39,0.39,0.39}
\definecolor{gray3}{rgb}{0.03,0.03,0.03}
\definecolor{gray40}{rgb}{0.40,0.40,0.40}
\definecolor{gray41}{rgb}{0.41,0.41,0.41}
\definecolor{gray42}{rgb}{0.42,0.42,0.42}
\definecolor{gray43}{rgb}{0.43,0.43,0.43}
\definecolor{gray44}{rgb}{0.44,0.44,0.44}
\definecolor{gray45}{rgb}{0.45,0.45,0.45}
\definecolor{gray46}{rgb}{0.46,0.46,0.46}
\definecolor{gray47}{rgb}{0.47,0.47,0.47}
\definecolor{gray48}{rgb}{0.48,0.48,0.48}
\definecolor{gray49}{rgb}{0.49,0.49,0.49}
\definecolor{gray4}{rgb}{0.04,0.04,0.04}
\definecolor{gray50}{rgb}{0.50,0.50,0.50}
\definecolor{gray51}{rgb}{0.51,0.51,0.51}
\definecolor{gray52}{rgb}{0.52,0.52,0.52}
\definecolor{gray53}{rgb}{0.53,0.53,0.53}
\definecolor{gray54}{rgb}{0.54,0.54,0.54}
\definecolor{gray55}{rgb}{0.55,0.55,0.55}
\definecolor{gray56}{rgb}{0.56,0.56,0.56}
\definecolor{gray57}{rgb}{0.57,0.57,0.57}
\definecolor{gray58}{rgb}{0.58,0.58,0.58}
\definecolor{gray59}{rgb}{0.59,0.59,0.59}
\definecolor{gray5}{rgb}{0.05,0.05,0.05}
\definecolor{gray60}{rgb}{0.60,0.60,0.60}
\definecolor{gray61}{rgb}{0.61,0.61,0.61}
\definecolor{gray62}{rgb}{0.62,0.62,0.62}
\definecolor{gray63}{rgb}{0.63,0.63,0.63}
\definecolor{gray64}{rgb}{0.64,0.64,0.64}
\definecolor{gray65}{rgb}{0.65,0.65,0.65}
\definecolor{gray66}{rgb}{0.66,0.66,0.66}
\definecolor{gray67}{rgb}{0.67,0.67,0.67}
\definecolor{gray68}{rgb}{0.68,0.68,0.68}
\definecolor{gray69}{rgb}{0.69,0.69,0.69}
\definecolor{gray6}{rgb}{0.06,0.06,0.06}
\definecolor{gray70}{rgb}{0.70,0.70,0.70}
\definecolor{gray71}{rgb}{0.71,0.71,0.71}
\definecolor{gray72}{rgb}{0.72,0.72,0.72}
\definecolor{gray73}{rgb}{0.73,0.73,0.73}
\definecolor{gray74}{rgb}{0.74,0.74,0.74}
\definecolor{gray75}{rgb}{0.75,0.75,0.75}
\definecolor{gray76}{rgb}{0.76,0.76,0.76}
\definecolor{gray77}{rgb}{0.77,0.77,0.77}
\definecolor{gray78}{rgb}{0.78,0.78,0.78}
\definecolor{gray79}{rgb}{0.79,0.79,0.79}
\definecolor{gray7}{rgb}{0.07,0.07,0.07}
\definecolor{gray80}{rgb}{0.80,0.80,0.80}
\definecolor{gray81}{rgb}{0.81,0.81,0.81}
\definecolor{gray82}{rgb}{0.82,0.82,0.82}
\definecolor{gray83}{rgb}{0.83,0.83,0.83}
\definecolor{gray84}{rgb}{0.84,0.84,0.84}
\definecolor{gray85}{rgb}{0.85,0.85,0.85}
\definecolor{gray86}{rgb}{0.86,0.86,0.86}
\definecolor{gray87}{rgb}{0.87,0.87,0.87}
\definecolor{gray88}{rgb}{0.88,0.88,0.88}
\definecolor{gray89}{rgb}{0.89,0.89,0.89}
\definecolor{gray8}{rgb}{0.08,0.08,0.08}
\definecolor{gray90}{rgb}{0.90,0.90,0.90}
\definecolor{gray91}{rgb}{0.91,0.91,0.91}
\definecolor{gray92}{rgb}{0.92,0.92,0.92}
\definecolor{gray93}{rgb}{0.93,0.93,0.93}
\definecolor{gray94}{rgb}{0.94,0.94,0.94}
\definecolor{gray95}{rgb}{0.95,0.95,0.95}
\definecolor{gray96}{rgb}{0.96,0.96,0.96}
\definecolor{gray97}{rgb}{0.97,0.97,0.97}
\definecolor{gray98}{rgb}{0.98,0.98,0.98}
\definecolor{gray99}{rgb}{0.99,0.99,0.99}
\definecolor{gray9}{rgb}{0.09,0.09,0.09}
\definecolor{gray}{rgb}{0.75,0.75,0.75}
\definecolor{green1}{rgb}{0.00,1.00,0.00}
\definecolor{green2}{rgb}{0.00,0.93,0.00}
\definecolor{green3}{rgb}{0.00,0.80,0.00}
\definecolor{green4}{rgb}{0.00,0.55,0.00}
\definecolor{greenyellow}{rgb}{0.68,1.00,0.18}
\definecolor{green}{rgb}{0.00,1.00,0.00}
\definecolor{grey0}{rgb}{0.00,0.00,0.00}
\definecolor{grey100}{rgb}{1.00,1.00,1.00}
\definecolor{grey10}{rgb}{0.10,0.10,0.10}
\definecolor{grey11}{rgb}{0.11,0.11,0.11}
\definecolor{grey12}{rgb}{0.12,0.12,0.12}
\definecolor{grey13}{rgb}{0.13,0.13,0.13}
\definecolor{grey14}{rgb}{0.14,0.14,0.14}
\definecolor{grey15}{rgb}{0.15,0.15,0.15}
\definecolor{grey16}{rgb}{0.16,0.16,0.16}
\definecolor{grey17}{rgb}{0.17,0.17,0.17}
\definecolor{grey18}{rgb}{0.18,0.18,0.18}
\definecolor{grey19}{rgb}{0.19,0.19,0.19}
\definecolor{grey1}{rgb}{0.01,0.01,0.01}
\definecolor{grey20}{rgb}{0.20,0.20,0.20}
\definecolor{grey21}{rgb}{0.21,0.21,0.21}
\definecolor{grey22}{rgb}{0.22,0.22,0.22}
\definecolor{grey23}{rgb}{0.23,0.23,0.23}
\definecolor{grey24}{rgb}{0.24,0.24,0.24}
\definecolor{grey25}{rgb}{0.25,0.25,0.25}
\definecolor{grey26}{rgb}{0.26,0.26,0.26}
\definecolor{grey27}{rgb}{0.27,0.27,0.27}
\definecolor{grey28}{rgb}{0.28,0.28,0.28}
\definecolor{grey29}{rgb}{0.29,0.29,0.29}
\definecolor{grey2}{rgb}{0.02,0.02,0.02}
\definecolor{grey30}{rgb}{0.30,0.30,0.30}
\definecolor{grey31}{rgb}{0.31,0.31,0.31}
\definecolor{grey32}{rgb}{0.32,0.32,0.32}
\definecolor{grey33}{rgb}{0.33,0.33,0.33}
\definecolor{grey34}{rgb}{0.34,0.34,0.34}
\definecolor{grey35}{rgb}{0.35,0.35,0.35}
\definecolor{grey36}{rgb}{0.36,0.36,0.36}
\definecolor{grey37}{rgb}{0.37,0.37,0.37}
\definecolor{grey38}{rgb}{0.38,0.38,0.38}
\definecolor{grey39}{rgb}{0.39,0.39,0.39}
\definecolor{grey3}{rgb}{0.03,0.03,0.03}
\definecolor{grey40}{rgb}{0.40,0.40,0.40}
\definecolor{grey41}{rgb}{0.41,0.41,0.41}
\definecolor{grey42}{rgb}{0.42,0.42,0.42}
\definecolor{grey43}{rgb}{0.43,0.43,0.43}
\definecolor{grey44}{rgb}{0.44,0.44,0.44}
\definecolor{grey45}{rgb}{0.45,0.45,0.45}
\definecolor{grey46}{rgb}{0.46,0.46,0.46}
\definecolor{grey47}{rgb}{0.47,0.47,0.47}
\definecolor{grey48}{rgb}{0.48,0.48,0.48}
\definecolor{grey49}{rgb}{0.49,0.49,0.49}
\definecolor{grey4}{rgb}{0.04,0.04,0.04}
\definecolor{grey50}{rgb}{0.50,0.50,0.50}
\definecolor{grey51}{rgb}{0.51,0.51,0.51}
\definecolor{grey52}{rgb}{0.52,0.52,0.52}
\definecolor{grey53}{rgb}{0.53,0.53,0.53}
\definecolor{grey54}{rgb}{0.54,0.54,0.54}
\definecolor{grey55}{rgb}{0.55,0.55,0.55}
\definecolor{grey56}{rgb}{0.56,0.56,0.56}
\definecolor{grey57}{rgb}{0.57,0.57,0.57}
\definecolor{grey58}{rgb}{0.58,0.58,0.58}
\definecolor{grey59}{rgb}{0.59,0.59,0.59}
\definecolor{grey5}{rgb}{0.05,0.05,0.05}
\definecolor{grey60}{rgb}{0.60,0.60,0.60}
\definecolor{grey61}{rgb}{0.61,0.61,0.61}
\definecolor{grey62}{rgb}{0.62,0.62,0.62}
\definecolor{grey63}{rgb}{0.63,0.63,0.63}
\definecolor{grey64}{rgb}{0.64,0.64,0.64}
\definecolor{grey65}{rgb}{0.65,0.65,0.65}
\definecolor{grey66}{rgb}{0.66,0.66,0.66}
\definecolor{grey67}{rgb}{0.67,0.67,0.67}
\definecolor{grey68}{rgb}{0.68,0.68,0.68}
\definecolor{grey69}{rgb}{0.69,0.69,0.69}
\definecolor{grey6}{rgb}{0.06,0.06,0.06}
\definecolor{grey70}{rgb}{0.70,0.70,0.70}
\definecolor{grey71}{rgb}{0.71,0.71,0.71}
\definecolor{grey72}{rgb}{0.72,0.72,0.72}
\definecolor{grey73}{rgb}{0.73,0.73,0.73}
\definecolor{grey74}{rgb}{0.74,0.74,0.74}
\definecolor{grey75}{rgb}{0.75,0.75,0.75}
\definecolor{grey76}{rgb}{0.76,0.76,0.76}
\definecolor{grey77}{rgb}{0.77,0.77,0.77}
\definecolor{grey78}{rgb}{0.78,0.78,0.78}
\definecolor{grey79}{rgb}{0.79,0.79,0.79}
\definecolor{grey7}{rgb}{0.07,0.07,0.07}
\definecolor{grey80}{rgb}{0.80,0.80,0.80}
\definecolor{grey81}{rgb}{0.81,0.81,0.81}
\definecolor{grey82}{rgb}{0.82,0.82,0.82}
\definecolor{grey83}{rgb}{0.83,0.83,0.83}
\definecolor{grey84}{rgb}{0.84,0.84,0.84}
\definecolor{grey85}{rgb}{0.85,0.85,0.85}
\definecolor{grey86}{rgb}{0.86,0.86,0.86}
\definecolor{grey87}{rgb}{0.87,0.87,0.87}
\definecolor{grey88}{rgb}{0.88,0.88,0.88}
\definecolor{grey89}{rgb}{0.89,0.89,0.89}
\definecolor{grey8}{rgb}{0.08,0.08,0.08}
\definecolor{grey90}{rgb}{0.90,0.90,0.90}
\definecolor{grey91}{rgb}{0.91,0.91,0.91}
\definecolor{grey92}{rgb}{0.92,0.92,0.92}
\definecolor{grey93}{rgb}{0.93,0.93,0.93}
\definecolor{grey94}{rgb}{0.94,0.94,0.94}
\definecolor{grey95}{rgb}{0.95,0.95,0.95}
\definecolor{grey96}{rgb}{0.96,0.96,0.96}
\definecolor{grey97}{rgb}{0.97,0.97,0.97}
\definecolor{grey98}{rgb}{0.98,0.98,0.98}
\definecolor{grey99}{rgb}{0.99,0.99,0.99}
\definecolor{grey9}{rgb}{0.09,0.09,0.09}
\definecolor{grey}{rgb}{0.75,0.75,0.75}
\definecolor{honeydew1}{rgb}{0.94,1.00,0.94}
\definecolor{honeydew2}{rgb}{0.88,0.93,0.88}
\definecolor{honeydew3}{rgb}{0.76,0.80,0.76}
\definecolor{honeydew4}{rgb}{0.51,0.55,0.51}
\definecolor{honeydew}{rgb}{0.94,1.00,0.94}
\definecolor{hotpink}{rgb}{1.00,0.41,0.71}
\definecolor{indianred}{rgb}{0.80,0.36,0.36}
\definecolor{ivory1}{rgb}{1.00,1.00,0.94}
\definecolor{ivory2}{rgb}{0.93,0.93,0.88}
\definecolor{ivory3}{rgb}{0.80,0.80,0.76}
\definecolor{ivory4}{rgb}{0.55,0.55,0.51}
\definecolor{ivory}{rgb}{1.00,1.00,0.94}
\definecolor{khaki1}{rgb}{1.00,0.96,0.56}
\definecolor{khaki2}{rgb}{0.93,0.90,0.52}
\definecolor{khaki3}{rgb}{0.80,0.78,0.45}
\definecolor{khaki4}{rgb}{0.55,0.53,0.31}
\definecolor{khaki}{rgb}{0.94,0.90,0.55}
\definecolor{lavenderblush}{rgb}{1.00,0.94,0.96}
\definecolor{lavender}{rgb}{0.90,0.90,0.98}
\definecolor{lawngreen}{rgb}{0.49,0.99,0.00}
\definecolor{lemonchiffon}{rgb}{1.00,0.98,0.80}
\definecolor{lightblue}{rgb}{0.68,0.85,0.90}
\definecolor{lightcoral}{rgb}{0.94,0.50,0.50}
\definecolor{lightcyan}{rgb}{0.88,1.00,1.00}
\definecolor{lightgoldenrod}{rgb}{0.93,0.87,0.51}
\definecolor{lightgoldenrod}{rgb}{0.98,0.98,0.82}
\definecolor{lightgray}{rgb}{0.83,0.83,0.83}
\definecolor{lightgreen}{rgb}{0.56,0.93,0.56}
\definecolor{lightgrey}{rgb}{0.83,0.83,0.83}
\definecolor{lightpink}{rgb}{1.00,0.71,0.76}
\definecolor{lightsalmon}{rgb}{1.00,0.63,0.48}
\definecolor{lightsea}{rgb}{0.13,0.70,0.67}
\definecolor{lightsky}{rgb}{0.53,0.81,0.98}
\definecolor{lightslate}{rgb}{0.47,0.53,0.60}
\definecolor{lightslate}{rgb}{0.47,0.53,0.60}
\definecolor{lightslate}{rgb}{0.52,0.44,1.00}
\definecolor{lightsteel}{rgb}{0.69,0.77,0.87}
\definecolor{lightyellow}{rgb}{1.00,1.00,0.88}
\definecolor{limegreen}{rgb}{0.20,0.80,0.20}
\definecolor{linen}{rgb}{0.98,0.94,0.90}
\definecolor{magenta1}{rgb}{1.00,0.00,1.00}
\definecolor{magenta2}{rgb}{0.93,0.00,0.93}
\definecolor{magenta3}{rgb}{0.80,0.00,0.80}
\definecolor{magenta4}{rgb}{0.55,0.00,0.55}
\definecolor{magenta}{rgb}{1.00,0.00,1.00}
\definecolor{maroon1}{rgb}{1.00,0.20,0.70}
\definecolor{maroon2}{rgb}{0.93,0.19,0.65}
\definecolor{maroon3}{rgb}{0.80,0.16,0.56}
\definecolor{maroon4}{rgb}{0.55,0.11,0.38}
\definecolor{maroon}{rgb}{0.69,0.19,0.38}
\definecolor{mediumaquamarine}{rgb}{0.40,0.80,0.67}
\definecolor{mediumblue}{rgb}{0.00,0.00,0.80}
\definecolor{mediumorchid}{rgb}{0.73,0.33,0.83}
\definecolor{mediumpurple}{rgb}{0.58,0.44,0.86}
\definecolor{mediumsea}{rgb}{0.24,0.70,0.44}
\definecolor{mediumslate}{rgb}{0.48,0.41,0.93}
\definecolor{mediumspring}{rgb}{0.00,0.98,0.60}
\definecolor{mediumturquoise}{rgb}{0.28,0.82,0.80}
\definecolor{mediumviolet}{rgb}{0.78,0.08,0.52}
\definecolor{midnightblue}{rgb}{0.10,0.10,0.44}
\definecolor{mintcream}{rgb}{0.96,1.00,0.98}
\definecolor{mistyrose}{rgb}{1.00,0.89,0.88}
\definecolor{moccasin}{rgb}{1.00,0.89,0.71}
\definecolor{navajowhite}{rgb}{1.00,0.87,0.68}
\definecolor{navyblue}{rgb}{0.00,0.00,0.50}
\definecolor{navy}{rgb}{0.00,0.00,0.50}
\definecolor{oldlace}{rgb}{0.99,0.96,0.90}
\definecolor{olivedrab}{rgb}{0.42,0.56,0.14}
\definecolor{orange1}{rgb}{1.00,0.65,0.00}
\definecolor{orange2}{rgb}{0.93,0.60,0.00}
\definecolor{orange3}{rgb}{0.80,0.52,0.00}
\definecolor{orange4}{rgb}{0.55,0.35,0.00}
\definecolor{orangered}{rgb}{1.00,0.27,0.00}
\definecolor{orange}{rgb}{1.00,0.65,0.00}
\definecolor{orchid1}{rgb}{1.00,0.51,0.98}
\definecolor{orchid2}{rgb}{0.93,0.48,0.91}
\definecolor{orchid3}{rgb}{0.80,0.41,0.79}
\definecolor{orchid4}{rgb}{0.55,0.28,0.54}
\definecolor{orchid}{rgb}{0.85,0.44,0.84}
\definecolor{palegoldenrod}{rgb}{0.93,0.91,0.67}
\definecolor{palegreen}{rgb}{0.60,0.98,0.60}
\definecolor{paleturquoise}{rgb}{0.69,0.93,0.93}
\definecolor{paleviolet}{rgb}{0.86,0.44,0.58}
\definecolor{papayawhip}{rgb}{1.00,0.94,0.84}
\definecolor{peachpuff}{rgb}{1.00,0.85,0.73}
\definecolor{peru}{rgb}{0.80,0.52,0.25}
\definecolor{pink1}{rgb}{1.00,0.71,0.77}
\definecolor{pink2}{rgb}{0.93,0.66,0.72}
\definecolor{pink3}{rgb}{0.80,0.57,0.62}
\definecolor{pink4}{rgb}{0.55,0.39,0.42}
\definecolor{pink}{rgb}{1.00,0.75,0.80}
\definecolor{plum1}{rgb}{1.00,0.73,1.00}
\definecolor{plum2}{rgb}{0.93,0.68,0.93}
\definecolor{plum3}{rgb}{0.80,0.59,0.80}
\definecolor{plum4}{rgb}{0.55,0.40,0.55}
\definecolor{plum}{rgb}{0.87,0.63,0.87}
\definecolor{powderblue}{rgb}{0.69,0.88,0.90}
\definecolor{purple1}{rgb}{0.61,0.19,1.00}
\definecolor{purple2}{rgb}{0.57,0.17,0.93}
\definecolor{purple3}{rgb}{0.49,0.15,0.80}
\definecolor{purple4}{rgb}{0.33,0.10,0.55}
\definecolor{purple}{rgb}{0.63,0.13,0.94}
\definecolor{red1}{rgb}{1.00,0.00,0.00}
\definecolor{red2}{rgb}{0.93,0.00,0.00}
\definecolor{red3}{rgb}{0.80,0.00,0.00}
\definecolor{red4}{rgb}{0.55,0.00,0.00}
\definecolor{red}{rgb}{1.00,0.00,0.00}
\definecolor{rosybrown}{rgb}{0.74,0.56,0.56}
\definecolor{royalblue}{rgb}{0.25,0.41,0.88}
\definecolor{saddlebrown}{rgb}{0.55,0.27,0.07}
\definecolor{salmon1}{rgb}{1.00,0.55,0.41}
\definecolor{salmon2}{rgb}{0.93,0.51,0.38}
\definecolor{salmon3}{rgb}{0.80,0.44,0.33}
\definecolor{salmon4}{rgb}{0.55,0.30,0.22}
\definecolor{salmon}{rgb}{0.98,0.50,0.45}
\definecolor{sandybrown}{rgb}{0.96,0.64,0.38}
\definecolor{seagreen}{rgb}{0.18,0.55,0.34}
\definecolor{seashell1}{rgb}{1.00,0.96,0.93}
\definecolor{seashell2}{rgb}{0.93,0.90,0.87}
\definecolor{seashell3}{rgb}{0.80,0.77,0.75}
\definecolor{seashell4}{rgb}{0.55,0.53,0.51}
\definecolor{seashell}{rgb}{1.00,0.96,0.93}
\definecolor{sienna1}{rgb}{1.00,0.51,0.28}
\definecolor{sienna2}{rgb}{0.93,0.47,0.26}
\definecolor{sienna3}{rgb}{0.80,0.41,0.22}
\definecolor{sienna4}{rgb}{0.55,0.28,0.15}
\definecolor{sienna}{rgb}{0.63,0.32,0.18}
\definecolor{skyblue}{rgb}{0.53,0.81,0.92}
\definecolor{slateblue}{rgb}{0.42,0.35,0.80}
\definecolor{slategray}{rgb}{0.44,0.50,0.56}
\definecolor{slategrey}{rgb}{0.44,0.50,0.56}
\definecolor{snow1}{rgb}{1.00,0.98,0.98}
\definecolor{snow2}{rgb}{0.93,0.91,0.91}
\definecolor{snow3}{rgb}{0.80,0.79,0.79}
\definecolor{snow4}{rgb}{0.55,0.54,0.54}
\definecolor{snow}{rgb}{1.00,0.98,0.98}
\definecolor{springgreen}{rgb}{0.00,1.00,0.50}
\definecolor{steelblue}{rgb}{0.27,0.51,0.71}
\definecolor{tan1}{rgb}{1.00,0.65,0.31}
\definecolor{tan2}{rgb}{0.93,0.60,0.29}
\definecolor{tan3}{rgb}{0.80,0.52,0.25}
\definecolor{tan4}{rgb}{0.55,0.35,0.17}
\definecolor{tan}{rgb}{0.82,0.71,0.55}
\definecolor{thistle1}{rgb}{1.00,0.88,1.00}
\definecolor{thistle2}{rgb}{0.93,0.82,0.93}
\definecolor{thistle3}{rgb}{0.80,0.71,0.80}
\definecolor{thistle4}{rgb}{0.55,0.48,0.55}
\definecolor{thistle}{rgb}{0.85,0.75,0.85}
\definecolor{tomato1}{rgb}{1.00,0.39,0.28}
\definecolor{tomato2}{rgb}{0.93,0.36,0.26}
\definecolor{tomato3}{rgb}{0.80,0.31,0.22}
\definecolor{tomato4}{rgb}{0.55,0.21,0.15}
\definecolor{tomato}{rgb}{1.00,0.39,0.28}
\definecolor{turquoise1}{rgb}{0.00,0.96,1.00}
\definecolor{turquoise2}{rgb}{0.00,0.90,0.93}
\definecolor{turquoise3}{rgb}{0.00,0.77,0.80}
\definecolor{turquoise4}{rgb}{0.00,0.53,0.55}
\definecolor{turquoise}{rgb}{0.25,0.88,0.82}
\definecolor{violetred}{rgb}{0.82,0.13,0.56}
\definecolor{violet}{rgb}{0.93,0.51,0.93}
\definecolor{wheat1}{rgb}{1.00,0.91,0.73}
\definecolor{wheat2}{rgb}{0.93,0.85,0.68}
\definecolor{wheat3}{rgb}{0.80,0.73,0.59}
\definecolor{wheat4}{rgb}{0.55,0.49,0.40}
\definecolor{wheat}{rgb}{0.96,0.87,0.70}
\definecolor{whitesmoke}{rgb}{0.96,0.96,0.96}
\definecolor{white}{rgb}{1.00,1.00,1.00}
\definecolor{yellow1}{rgb}{1.00,1.00,0.00}
\definecolor{yellow2}{rgb}{0.93,0.93,0.00}
\definecolor{yellow3}{rgb}{0.80,0.80,0.00}
\definecolor{yellow4}{rgb}{0.55,0.55,0.00}
\definecolor{yellowgreen}{rgb}{0.60,0.80,0.20}
\definecolor{yellow}{rgb}{1.00,1.00,0.00}

\usepackage{comment}
\usepackage{amssymb}
\usepackage{amsfonts}

\usepackage{subcaption}
\usepackage{algorithmic}
\usepackage{algorithm}
\usepackage{pifont}
\usepackage{booktabs}

\usepackage{lipsum}
\usepackage{array}
\newcolumntype{P}[1]{>{\centering\arraybackslash}p{#1}}

\newcommand{\LP}[1]{\textcolor{red}{}}

\newcommand{\longonly}[1]{}

\usepackage{adjustbox}

\newcolumntype{R}[2]{%
    >{\adjustbox{angle=#1,lap=\width-(#2)}\bgroup}%
    l%
    <{\egroup}%
}

\graphicspath{{figures/},{figures/new_diagrams/}}

\usepackage{hyperref}

\allowdisplaybreaks[2]
\renewcommand{\baselinestretch}{0.99}
\usepackage{mdwlist}

\let\itemize\undefined
\makecompactlist{itemize}{stditemize}

\let\enumerate\undefined
\makecompactlist{enumerate}{stdenumerate}

\usepackage[tracking=false,kerning=true,spacing=true]{microtype}
\title{\LARGE \bf

Cross-Modal Contrastive Learning of Representations
for Navigation using Lightweight, Low-Cost Millimeter Wave Radar for Adverse Environmental Conditions\\
}

\author{Jui-Te Huang, Chen-Lung Lu, Po-Kai Chang, Ching-I Huang, Chao-Chun Hsu, \\
Zu Lin Ewe, 
Po-Jui Huang, and Hsueh-Cheng Wang$^{*}$
\thanks{All authors are affiliated with National Chiao
Tung University, Taiwan. Corresponding author email:
        {\tt\small hchengwang@g2.nctu.edu.tw}}%
}

\begin{document}


\maketitle

\begin{abstract}

Deep reinforcement learning (RL), where the agent learns from mistakes, has been successfully applied to a variety of tasks. With the aim of learning collision-free policies for unmanned vehicles, deep RL has been used for training with various types of data, such as colored images, depth images, and LiDAR point clouds, without the use of classic map--localize--plan approaches.
However, existing methods are limited by their reliance on cameras and LiDAR devices, which have degraded sensing under adverse environmental conditions (e.g., smoky environments). 
In response, we propose the use of single-chip millimeter-wave (mmWave) radar, which is lightweight and inexpensive, for learning-based autonomous navigation. However, because mmWave radar signals are often noisy and sparse, we propose a cross-modal contrastive learning for representation (CM-CLR) method that maximizes the agreement between mmWave radar data and LiDAR data in the training stage. We evaluated our method in real-world robot compared with 1) a method with two separate networks using cross-modal generative reconstruction and an RL policy and 2) a baseline RL policy without cross-modal representation. Our proposed end-to-end deep RL policy with contrastive learning successfully navigated the robot through smoke-filled maze environments and achieved better performance compared with generative reconstruction methods, in which noisy artifact walls or obstacles were produced. All pretrained models and hardware settings are open access for reproducing this study and can be obtained at  \url{https://arg-nctu.github.io/projects/deeprl-mmWave.html}.

\end{abstract}

\section{Introduction}
\label{sec:Introduction}

Navigation and collision avoidance are fundamental capabilities for mobile robots, and state estimation and path planning, which are integral to such capabilities, have undergone much development. Deep reinforcement learning (RL) is another method for realizing these capabilities, where robots learn to solve problems through trial and error. Extensive efforts have been made to formulate training methods for deep RL models using various types of data, such as those from RGB cameras, depth images~\cite{savva-habitat19iccv, zhang2017deep}, and laser range finder (LiDAR) point clouds~\cite{pfeiffer2017perception, pfeiffer2018reinforced}. However, optical sensors are often useless in adverse weather and environmental conditions: LiDAR inputs are degraded by rain, fog, and dust, and cameras are sensitive to variable lighting in dark and sunny weather. In search and rescue missions, for example, adverse environmental conditions are commonly encountered and result in degraded sensing. In the DARPA Subterranean Tunnel Circuit, there are challenging environments in unknown and long coal mine tunnels with smoke filled adverse conditions that impair commonly used camera and LiDAR sensors.

\begin{figure}[t]
\includegraphics[width=1.0\columnwidth]{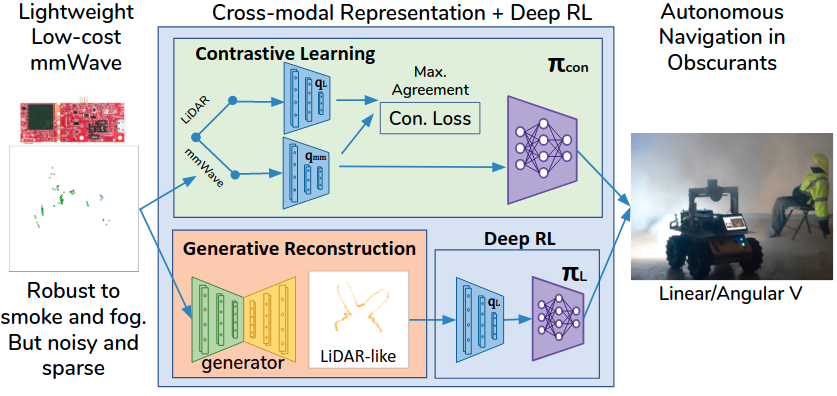}
\centering

\caption{Cross-modal contrastive learning of representations is proposed to generate representations of range data from mmWave radars, which can operate even in smoke-filled environments. The proposed method is compared with a generative reconstruction model.}

\label{fig:teaser}
\end{figure}

Millimeter wave (mmWave) radar is an alternative sensor that has promise for overcoming these challenges that can operate despite variable weather and lighting conditions \cite{cen2019radar, cen2018precise}. 
Recently, low-cost, low-power, and lightweight single-chip mmWave radar has emerged as an innovative sensor modality. Methods are proposed to provide state estimation \cite{almalioglu2020milli} through neural networks or reconstruct dense grid maps \cite{lu2020millimap} with conditional GAN.
Although the results show promise, navigation using maps reconstructed from mmWave data faces problems from sensor noise and multipath reflections, which may be mistakenly rendered as obstacles on the reconstructed map. Another thought is to have the RL agent learn the noise model of mmWave radars and correspond accordingly. However, directly training an end-to-end deep RL policy network in a simulation using mmWave data requires some channel models~\cite{hemadeh2017millimeter} that are not yet available in commonly used simulation frameworks in the robotics literature. Therefore, a large amount of mmWave training data from the real world for offline RL training is needed which makes the method inefficient to almost infeasible.

To this extent, we explored contrastive learning to learn cross-modal representations between mmWave and LiDAR. Cross-modal representations were analyzed in \cite{bonatti2019learning}, where the authors proposed a variational autoencoder (VAE)\textendash based method, which was to be applied in real-world aerial navigation, for learning robust visuomotor policies from simulated data alone. 
In our study, we leveraged the contrastive learning framework by regarding mmWave radar data as an augmentation of LiDAR data. Specifically, an encoder was trained to learn a sufficient number of patterns in mmWave data for the RL agent, which then navigated an unmanned ground vehicle (UGV) using only noisy raw radar inputs with no reconstruction.

We used lightweight and low-cost single-chip mmWave sensors, which can operate under challenging foggy or smoke-filled environments. To enable navigation despite noisy and low-resolution mmWave radar inputs, we proposed approaches for learning how to control UGVs and evaluated these approaches in a smoke-filled environment. We summarize our contributions as follows:

\begin{enumerate}
    \item \textbf{We show how cross-modal contrastive learning of representations (CM-CLR) can be used to maximize the agreement between mmWave and LiDAR inputs}:
    We introduced contrastive loss into our network and formulated a training method for achieving consistency between mmWave and LiDAR latent representations during forward propagation. We collected a dataset comprising both mmWave and LiDAR inputs in various indoor environments, such as corridors, open areas, and parking lots. All LiDAR data were only used for training.

    \item \textbf{We demonstrate that deep RL with cross-modal representations can be used for navigation}:
     The representation obtained from contrastive learning was integrated into an end-to-end deep RL network $\pi_{con}$ (Fig.~\ref{fig:teaser}). We demonstrated that inputs from only mmWave radar signals enabled navigation, unlike LiDAR and camera signals, in visually challenging smoke-filled maze environments. 
    
    \item \textbf{We evaluated our approaches in real-world smoke-filled environments}:
    We comprehensively evaluated a) the proposed end-to-end contrastive learning method and b) deep generative models that generate LiDAR-like range data as\ inputs for a deep RL\textendash based control policy $\pi_L$ network. We used a conditional generative adversarial network (cGAN) and VAEs, where raw mmWave inputs are used to reconstruct dense LiDAR-like range data. We showed that the navigation achieved by our proposed method can better avoid obstacles and respond to situations in which the vehicle is trapped.

\end{enumerate}

\section{Related Work}
\label{sec:related}

\subsection{Contrastive Learning}

Contrastive learning is a type of learning framework where the model learns patterns in a dataset, which are organized in terms of similar versus dissimilar pairs, while subject to similarity constraints.
In \cite{chen2020simple}, a simple contrastive learning framework (SimCLR) was proposed for learning visual representations, with the application of simple data augmentation for contrastive prediction tasks, and the authors incorporated cosine similarity temperature-scaled cross-entropy loss into their method. The performance of the proposed framework  matched that of supervised learning methods. That study further analyzed the effects of different data augmentation operators and concluded that a combination of cropping and color distortions maximizes learning efficiency. Another study \cite{laskin_srinivas2020curl} proposed the CURL framework for learning with high-dimensional inputs (i.e., images), in which contrastive learning, for representation learning, is combined with RL, for control policy learning. The framework outperformed state-of-the-art image-based RL in several DeepMind control suite tasks and Atari game tasks and even nearly matched the sample efficiency of using state-based features for RL. In this work, we considered mmWave data to be an augmentation of LiDAR data applied a contrastive learning framework to learn cross-modal representations through maximization of the agreement between encoded mmWave radar data and LiDAR data. 

\subsection{Deep Reinforcement Learning for Navigation}
Recent studies have trained deep networks either through a pre-generated occupancy map for motion commands or a mapless end-to-end approach that relates range inputs to actions. Kahn \textit{et al}. \cite{kahn2020badgr} demonstrated affordance prediction for future paths with a network trained through self-supervised learning, which allowed for navigation in paths that were previously regarded as untraversable in the classic SLAM and planning approach. Niroui \textit{et al}. \cite{niroui2019deep} presented a map-based frontier prediction network that was trained using the RL framework in urban search and rescue applications. Both of these works combined learning modules with classic control modules, in which deep networks are not responsible for predicting motion commands. On the other hand, \cite{hu2020voroni} proposed a method for multi-robot navigation with a high-level voroni-based planar and a low-level deep RL-based controller, where the RL agent is only responsible for the controller. Mapless end-to-end approaches (e.g., in\cite{pfeiffer2017perception, pfeiffer2018reinforced, tai2017virtual}) use range data as inputs to predict suitable motion commands through a trained network. Pfeiffer \textit{et al}. \cite{pfeiffer2017perception} developed a network that learns through feature extraction in a convolutional neural network (CNN) layer and through decision making in fully connected (FC) layers using data-driven behavioral cloning. In another study \cite{tai2017virtual}, an agent network was trained through RL algorithms \cite{lillicrap-2015-ddpg, gu2017deep} in virtual environments. Another study \cite{pfeiffer2018reinforced} pretrained the agent network using behavioral cloning prior to training with the RL framework. Finally, in two studies\cite{pfeiffer2018reinforced, tai2017virtual}, further downsampling was performed or fewer range data points were used to avoid overfitting or  obtain better sim-to-real performance. 

However, all these studies used LiDAR devices or cameras as input sensors, which are vulnerable in obscurant-filled environments. There has been earlier work using an array of sonars~\cite{walter1987sonar, fazli2005wall}, which used the propagation of acoustic energy and could penetrate smoke to sense obstacles from the environment~\cite{Kleeman2008}. However, low angular resolution and short range limited the applications.
Thus, we used mmWave radar instead of LiDAR for range sensing because mmWave signals can penetrate smoke and fog. Our network architecture is similar to that proposed in \cite{pfeiffer2017perception}, which included feature extraction CNN layers and decision-making FC layers.

\section{Deep Reinforcement Learning with Cross-Modal Contrastive Learning} \label{sec-method}

\subsection{Pre-processing}

We projected all range data to a two-dimensional (2D) plane with a 240$^{\circ}$ front-facing field of view and 1$^{\circ}$ resolution. Our range data were obtained from LiDAR as $x_l \in \mathbb{R}^{241}$ and mmWave as $x_m \in \mathbb{R}^{241}$ at each time step. Each of the four mmWave radars were running at 15 Hz, and synchronized into 10 Hz. We then applied an outlier removal filter by clustering the point clouds before projection. To do so, we first accumulated 5 consecutive frames to increase features. Then each point would be kept if there exist 3 neighbor points within 1.2 meter radius. There may exist missing value at a certain angles after applying the filtering, and we applied a padding of a maximum range value of 5 m to ensure sufficient input dimensions.




\begin{figure}[t]
    \centering
    \begin{subfigure}[b]{0.48\columnwidth}
        \centering
        \includegraphics[width=\columnwidth]{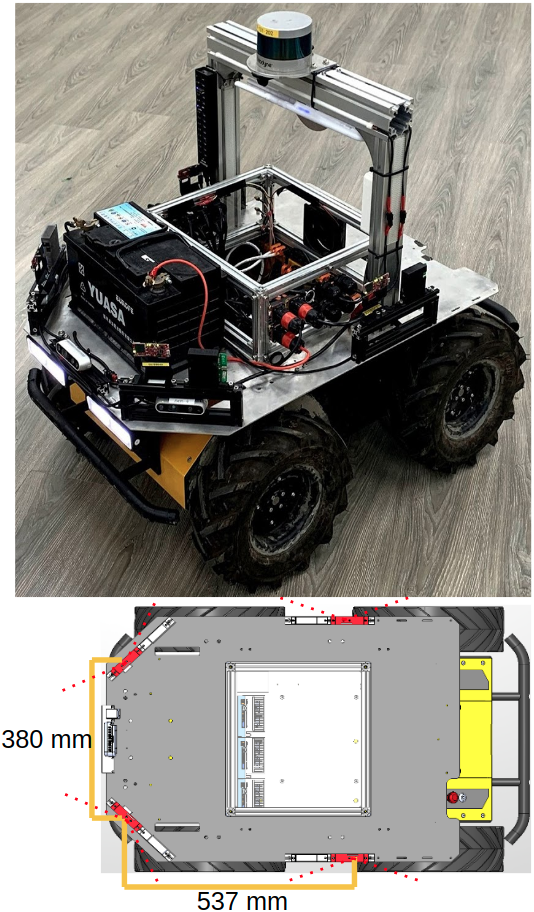}
        \caption{Sensor configuration 1.}
    \end{subfigure}%
    ~ 
    \begin{subfigure}[b]{0.48\columnwidth}
        \centering
        \includegraphics[width=\columnwidth]{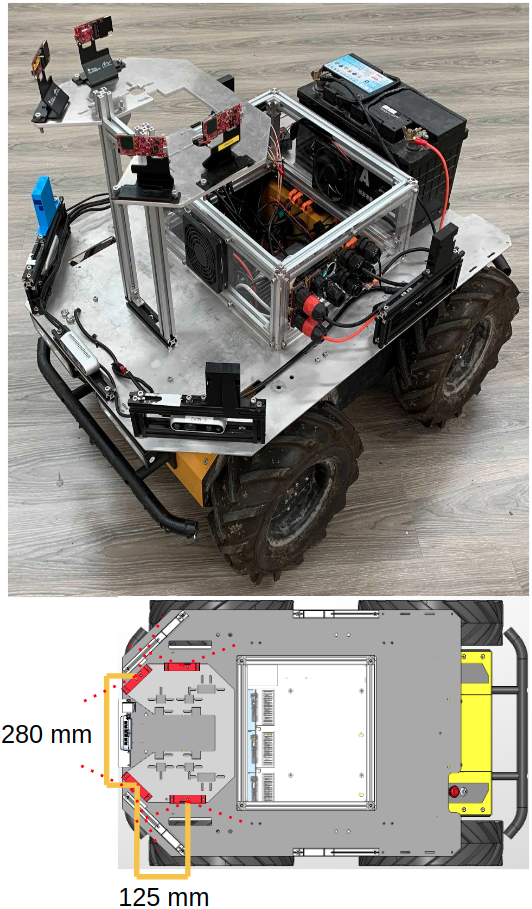}
        \caption{Sensor configuration 2.}
    \end{subfigure}
    \caption{Two configurations of mmWave radar modules placements mounted on a Clearpath Husky. The field of views of both configurations were constrained to 240$^{\circ}$. In (a) the Velodyne 3D LiDAR was only for data collection. All experiments and data collection in the paper were conducted using configuration 1. (b) Configuration 2 was designed for the potential use of smaller robots, and the discussions are described in Sec.\ref{sec:general_discuss}}
    \label{fig:hardware}
\end{figure}

\subsection{Contrastive Learning}
\begin{figure}[t]
    \includegraphics[width=\columnwidth]{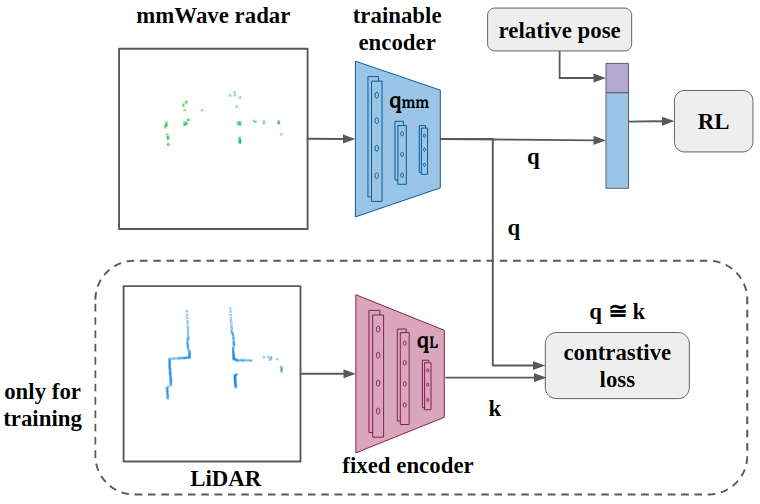}
    \caption{Query encoder for mmWave radar range data $q_{mm}$ is used to generate a representation $q$ that is similar to the representation $k$ generated by the key encoder $q_L$, which are trained from LiDAR range data in simulation. The weights of $q_L$ are set unchanged during contrastive learning.}
    \label{fig:network}
\end{figure}

To overcome the problem of noisy and sparse mmWave radar signals in navigation, we proposed contrastive learning, the agreement between mmWave and LiDAR inputs is maximized through the tuning of a trainable encoder with contrastive loss.  

Intuitively, learning a control policy can be done directly from mmWave-obtained range data, without the use of LiDAR data. However, there exist two issues of noise and lack of simulator access. Training using real-world data, which lack false cases (e.g., running into walls), results in agents that have a high frequency of collisions with obstacles. In contrastive learning, the agent learns a good representation of some high-dimensional inputs through different forms of data augmentation and contrastive loss. We regarded mmWave range data as a noisy and randomly dropped augmentation of LiDAR range data, and we trained an encoder using mmWave data to generate a latent representation that is identical to that from the LiDAR ground truth. Because the agent in the simulation has learnt a good navigation strategy with such a representation, our new agent should, in theory, be capable of navigating highly similar representations that are generated from the trained encoder and mmWave input.

\subsection{Learning Using LiDAR Data for Navigation in a Virtual Subterranean Environment}

First, we aimed to train the agent to navigate in long corridors, tunnels, or caves in subterranean environments using RL algorithms. 
The reward settings were expected to encourage the agent to move forward and avoid collisions at the same time to obtain high rewards. The agent should also receive some rewards to keep moving and exploring. High angular velocity was expected to decrease the rewards so that the agent will not make sharp turns unless there was no opening sensed at front. The model was used as a low-level collision avoidance controller instead of a high-level planar.

To do so, we used a variety of virtual subterranean environments~\cite{virtual-subt} in Gazebo simulation software, comprising an artificial tunnel with obstacles and cliffs, a subway station, and a natural cave. We used a cave environment because it has the most diverse sizes and shapes of tunnels and ramps. 
We followed the OpenAI Gym coding style to integrate our gym environment with Gazebo, analyze data on the agent's observations, calculate rewards, assign actions, and reset the agent. The settings and justifications for them are as follows:
\begin{itemize}
    
    \item Observations: The three-dimensional simulated LiDAR inputs constituted an observation space. As stated in Section \ref{sec-method}, we sampled the range data for $-120^{\circ}$ to $120^{\circ}$, with a $1^{\circ}$ resolution from the point cloud of a height reaching the UGV's height, resulting in a total of 241 range values. The agent is also given the information of motion direction, the relative 2D position of the agent is defined with respect to the previous time point by $x_p \in \mathbb{R}^{2}$. The total observation $O$ of our RL agent is the combination of $x_l$ and $x_p$, which has the dimension of $O \in \mathbb{R}^{243}$.
    \item Actions: A\ continuous space of linear and angular velocities was used to represent actions. The linear velocity space was normalized to $[0, 1]$, and the angular velocity space was normalized to $[-1, 1]$. Angular velocity was represented as a continuous rather than discrete space to allow for smooth motion (which is desirable), and linear velocity was represented to prevent the agent from moving backward.
    \item Reward: To encourage the agent to navigate indefinitely, we designed the rewards according to three preference. The first, \(r\_action\), incentivizes the agent to move straight and perform as few turns as possible. The second, \(r\_move\), incentivizes the agent to keep moving. Finally, the third, \(r\_collision\), disincentivizes the agent from colliding into obstacles, which results in the end of an episode. Specifically, these three rewards and their combination were designed as follows.
    \begin{flalign*}
        & r\_action = 2^{(1-|\omega|)\times5} \text{, where } \omega \text{ is the angular velocity}  \\
        & r\_move = 
            \begin{cases}
                10\quad{if}\quad\text{the distance} \\ 
                \quad\quad\quad\quad\text{between frames} > 0.04m\\
                -10\quad{else}
            \end{cases} \\
        & r\_collision = 
            \begin{cases}
                -50\quad{if}\quad\text{collision} \\ 
                0\quad{else}
            \end{cases} \\
        & r = \lambda \times r\_action + r\_move + r\_collision
    \end{flalign*}
    The reward value that was sent to the agent was converted using a log scale: \(r\_agent=sign(r)\times log(1+|r|)\).
\end{itemize}

The deep deterministic policy gradient (DDPG) \cite{lillicrap-2015-ddpg} is an off-policy RL algorithm that trains a deterministic policy to be the agent's control policy. The DDPG uses an actor-critic method to learn a deterministic policy $\pi^\theta$ with parameter $\theta$ and Q-value estimator $Q^\phi$ with parameter $\phi$. In contrast to the DDPG, where the network is updated using transitions $t=(o_{t}, a, r, o_{t+1})$ sampled from the experience replay buffer, the recurrent deterministic policy gradient (RDPG) algorithm \cite{Heess2015MemorybasedCW} uses the recurrent network and trains the network with an entire trajectory history \(h_{t}=(o_{1},a_{1},o_{2},a_{2},\dots,a_{t-1},o_{t})\) and back propagation through time. 
The critic network updates $\phi$ to minimize the Bellman error, and
the actor network's objective is to maximize $Q^\phi(h_t,a_t)$.

We used DDPG and RDPG algorithms to train collision avoidance policies with data collected in the Gazebo virtual subterranean environment. Our neural network architecture comprised minimum pooling for downsampling the inputs, two convolution layers for feature extraction, two FC layers, one recurrent layer with an LSTM cell, and one FC layer to generate action outputs. The network architecture was similar to~those of \cite{pfeiffer2017perception} and \cite{fan2018fully}, which used three consecutive frames, whereas our network includes LSTM cells. Studies have reported the importance of including a recurrent network when learning sequential actions. Training process took 55 hours for 4 million steps on an Intel i7-8700 CPU with Nvidia RTX2070 GPU. We denoted our control policy trained in the simulation using LiDAR data as $\pi_L$.

\subsection{Navigation With Policy From Contrastive Learning}

To calculate the contrastive loss, we separated the original RL network into a feature extraction network, which comprised the convolution layers, and a decision-making network, which comprised the FC layers and the recurrent layer. The goal of the contrastive learning was to maximize the similarity of the encoded representations between the mmWave input and the matching ground-truth LiDAR input; this was done to allow the mmWave representation to be used by the decision-making network, which was trained with RL in the simulation. Thus, the key encoder $q_{L}(k|x_l)$  and the query encoder $q_{mm}(q|x_m)$ of contrastive learning were designed as the architecture of the feature extraction layers of the RL policy network. During contrastive learning, real-world LiDAR range data were used as the input for the key encoder. The mmWave range data, which can be construed as an augmentation of ground truth LiDAR data, were taken as the input of the query encoder.
 In CURL\cite{laskin_srinivas2020curl}, the bilinear inner product with a momentum key encoder was used for training with contrastive loss. However, we did not want to change the key encoder, which was already trained with RL during simulation. We used a fixed key encoder with euclidean distances to measure the similarity between $k$ and $q$.
 \[
\mathcal{L}_{contrastive}= \mathbb{E}_{x_l,x_m}[(q_{mm}(q|x_m)-q_{L}(k|x_l))^2]
\]

We trained the network with the collected dataset to converge (20 epochs), and stopped the training to prevent overfitting. The performance of the trained mmWave control policy $\pi_{con}$ was evaluated in a real-world maze.

\section{Generative Reconstruction Baselines}

We considered generative reconstructions of mmWave range data into LiDAR-like data to serve as a key baseline because prior work \cite{lu2020millimap} showed promising results. We implemented two state-of-the-art supervised deep generative models, which were later combined with control policies for navigation tasks.

\subsection{Conditional Generative Adversarial Network for Range Data}

A cGAN comprises a generator $G$ and a discriminator $D$. Given an mmWave range data point $x_m$ and a random vector $z$ as the input, the generator attempts to generate a much clearer and denser range data point $y$ that is similar to the ground truth LiDAR range data point $x_l$. By contrast, the discriminator attempts to distinguish the real LiDAR range data point $x_l$ generated from $y$. The objective of a cGAN can be expressed as
follows. 
\begin{multline*}
\mathcal{L}_{cGAN}(G,D)=\mathbb{E}_{x_m,y}[logD(x_m,y)]+ \\
\mathbb{E}_{x_m,z}[log(1-D(x_m,G(x_m,z)))]
\end{multline*}

To make the generated range data even more similar to the LiDAR ground truth, we adapted L1 loss in the objective function following the pix2pix approach of a previous work\cite{Isola_2017_CVPR}.
\[
\mathcal{L}_{L1}(G)=\mathbb{E}_{x_m,y,z}[ |y-G(x_m,z)| ]
\]
The final objective function for the generator is
as follows.
\[
\mathcal{L}_{total} = \mathcal{L}_{cGAN} + \lambda_{1}\mathcal{L}_{L1}
\]

We designed the generator to have an encoder--decoder structure. The encoder has the same convolution feature extraction architecture as that of our deep RL networks. For the decoder model, we implemented three deconvolution layers and an FC layer to adjust the output dimension.
For our discriminator, we used the Markovian discriminator technique\cite{LiW-2016-markovian}, also called patchGAN in pix2pix\cite{Isola_2017_CVPR}. The patched discriminator has four layers of one-dimensional convolution for feature extraction, and the output is a $1 \times 14$ patch. We also tested a nonpatched discriminator that outputs a single value to discriminate real objects from counterfeit ones. We found that the nonpatched discriminator had slightly better L1 performance. 
We trained the model using our collected dataset. The data collection process is described in Section \ref{sec-data-collection}.

\subsection{Variational Autoencoder}
In addition to a cGAN, we used a similar network architecture to train a VAE \cite{Kingma2014AutoEncodingVB}, which
comprised a probabilistic encoder 
\(q(z|m)\) and a generative model \(p(l|z)\), where $z$ is the latent variable, sampled from the output of the encoder. 
The objective function of the VAE minimizes the variational lower bound of the log likelihood. In \cite{spurr2018cvpr}, the objective function was interpreted as accounting for probabilities across data modalities. The objective function in our case is as follows.
\[
 \mathbb{E}_{z \sim q(z|x_m)}\left[logp(x_l|z)-D_{KL}(q(z|x_m)||p(z))\right]
\]

\section{Evaluation} \label{Experiments}

\subsection{Data Collection} \label{sec-data-collection}
Range data from four 60-GHz TI 6843 mmWave radar modules and a Velodyne VLP-16 LiDAR and motion commands were recorded at a frame rate of 5 Hz as training data using a Clearpath Husky. A total of 110\thinspace381 frames (approximately 6 robot hours) were recorded. The experimental environments included a corridor, a parking area, and outdoor open areas. The indoor corridors had widths from 2 to 4 m; the parking area had a wide passage with a maximum width of 6 m, and the outdoor open area had a maximum width of 10 m. All data in this study, which comprised the aforementioned types of data as well as images and wheel odometry, are open source and in ROS bag and hdf5 file formats.

\begin{figure}[ht]
    \includegraphics[width=\columnwidth]{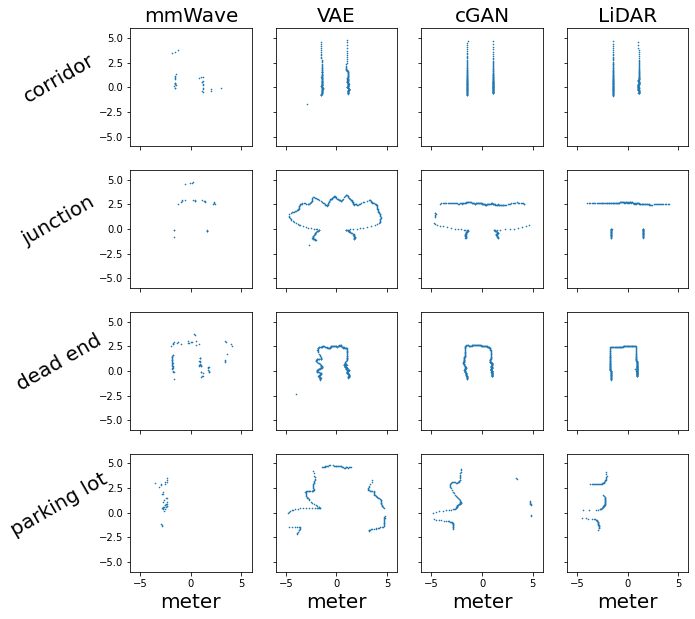}
    \caption{Raw mmWave range data and ground truth LiDAR range data collected in different environments with different reconstruction methods.}
    \label{fig:mm-vae-cgan}
\end{figure}

\begin{figure*}[ht]
    \centering
    \begin{subfigure}[b]{0.5\textwidth}
        \centering
        \includegraphics[width=\columnwidth]{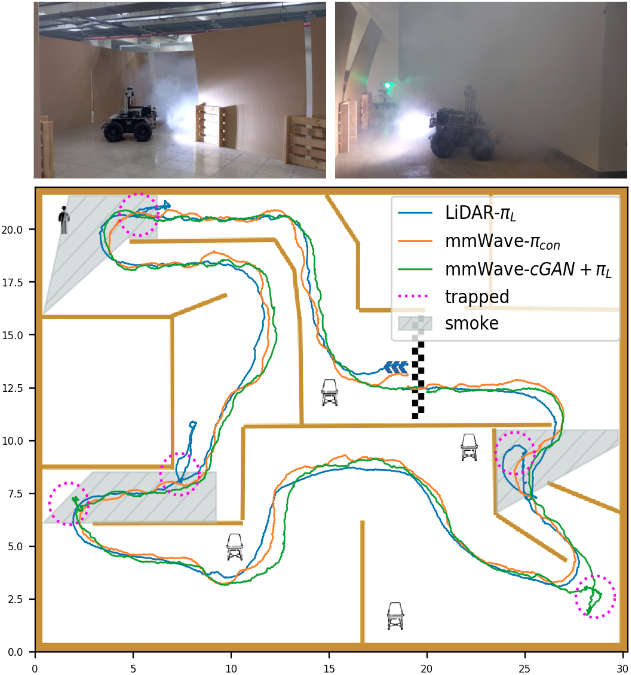}
        \caption{A maze environment built by cardboard.}
    \end{subfigure}%
    ~ 
    \begin{subfigure}[b]{0.5\textwidth}
        \centering
        \includegraphics[width=\columnwidth]{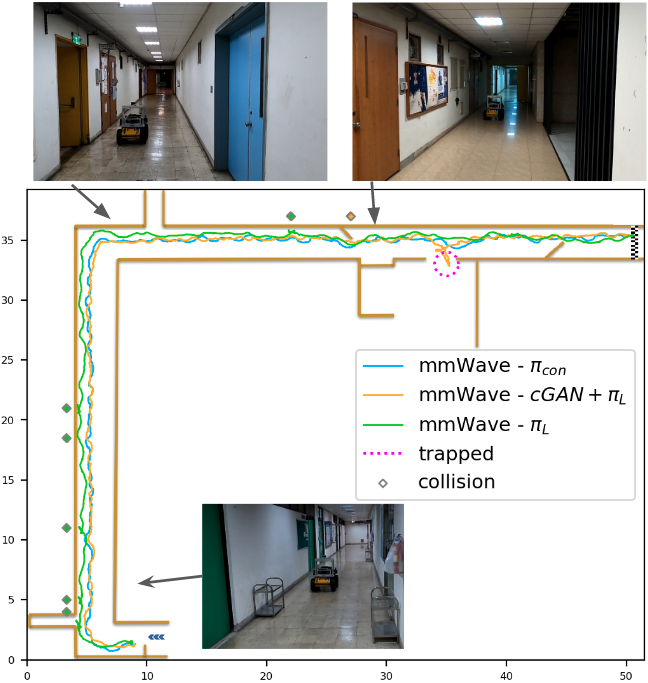}
        \caption{A general indoor environment.}
    \end{subfigure}
    \caption{Navigation experiments and visualized trajectories of various control policies. (a) A maze environment built by cardboard. The robot was trapped due to the smoke (LiDAR + $\pi_L$) or phantom walls (mmWave-$cGAN+\pi_L$). (b) 
    A general indoor environment, where mmWave-$cGAN+\pi_L$ has some cases of collisions and instances of being trapped caused by reconstruction errors.}
    \label{fig:exp-maze}
    \vspace{-16pt}
\end{figure*}

\subsection{Reconstruction Results}
In addition to the previously mentioned reconstruction methods, the polynomial curve fitting method in the polar coordinate was used as a baseline. Among the 1st-order to 15th-order curves, the 8th-order curve had the lowest L1 distance value. To evaluate the performance of different reconstruction methods, we compared the L1 distance between the reconstructed mmWave data and the ground truth LiDAR data. The results are displayed in Table~\ref{table:comparison}. The  qualitative results of different reconstruction methods are illustrated in Fig. \ref{fig:mm-vae-cgan}. Data collected in scenes that were from a corridor or parking lot and were not a maze used in the training steps were all considered in the evaluation.
\vspace{-6pt}
\begin{table}[ht]
\centering
\footnotesize
\caption{L1 loss of various reconstruction methods in different environments.}

\begin{tabular}{cccc}
\hline
    Method        & Corridor  &  Parking Lot  & Maze  \\ [4pt]\hline
    raw mmWave    & 1.99    &  0.39       & 2.53      \\ 
    curve fitting & 0.75    &  1.52       & 1.84      \\ 
    VAE           & 0.23    & 0.27        & 1.65      \\ 
    cGAN          & \bf0.09 & \bf0.11     & \bf1.53      \\ \hline
\end{tabular}
\vspace{-8pt}
\label{table:comparison}
\footnotesize
\end{table}
 As indicated in Table~\ref{table:comparison}, the raw mmWave data had a lower L1 distance in the parking lot than in the other environments. This is because the parking lot had wider passages, which caused most range data points to exceed the stipulated maximum distance of 5 m. Thus, most mmWave data and LiDAR ground truth data points were forced to be at 5 m, which resulted in a low L1 distance, regardless of whether the value was absent or higher. Among all the reconstruction methods, the cGAN had the lowest L1 distance, which implies that the reconstruction data of the cGAN were most similar to the ground truth LiDAR data. Both the cGAN and VAE could generate geometries that were similar to the ground truth, but the cGAN outperformed the VAE by generating straighter, smoother, and more consistent walls and fewer ``phantom walls." We found that both methods had difficulty reconstructing the exact shapes of cars in the parking lot scene using sparse mmWave data. By contrast, reconstructing the geometry of corridors was easy, as indicated by the lower L1 loss in the corridor scene.
We noted that the average performance decreased in our test scene in the maze of the underground basement. This might be because the mmWave signals could penetrate the cardboard, resulting in inaccurate reconstructions.

\subsection{Navigation in a Smoke-Filled Maze}

\subsubsection{Experiment Environment}
The testing environment was a cardboard maze in a basement (23 m $\times$ 28 m) with widths ranging from 2.25 m to 6 m, and the walls were 1.78 m high. One lap was approximately 100 m long and contained 20 turns, which were mostly $90^{\circ}$ turns, except for some sharp turns. We used four chairs and one mannequin as obstacles. The robot navigated in the maze, similar to that in the tunnel without intersections.
The LeGO-LOAM~method\cite{shan2018lego} was applied with the LiDAR data to obtain robot trajectories; the trajectories of some methods are plotted in Fig. \ref{fig:exp-maze} for qualitative evaluation. 

\subsubsection{Metrics}

Robot navigation failures were quantified according to the number of \textbf{collisions} and the number of instances of being \textbf{trapped}, which were a result of noisiness or sparseness in the mmWave radar data. Specifically, the robot collided into an obstacle when the obstacle was undetected and became trapped when the range estimates were inaccurate, which could be due, for example, to phantom walls. We also recorded odometry results from whee and LiDAR, but the distance traveled did not significantly differ between the methods evaluated.

\subsubsection{Ground truth}
We used navigation with LiDAR data, using both the RL control policy $\pi_{L}$ and classic A* planning $\pi_{cl}$ in a smoke-free environment, as the ground truth. The goal points of classic planning were set to follow the walls, and a pure pursuit algorithm and PID controller were used. Both methods resulted in the robot safely traversing the maze without colliding or being trapped. 

\subsubsection{Smoke trials}

A smoke machine emitted dense smoke at three locations (Fig.~\ref{fig:exp-maze}) for 15s before the UGV arrived. To adhere to safety regulations, we did not fill the entire basement with smoke. We conducted a few smoke trials for $\pi_L$, $cGAN+\pi_L$, and $\pi_{con}$ to observe how the smoke affected the control policies. The smoke prevented the LiDAR signals from penetrating the passage, which caused the UGV  to be trapped. We found that navigation methods using mmWave radar were unaffected by smoke.

\vspace{-8pt}
\begin{table}[ht]
\centering
\caption{Navigation performance in maze with respect to number of collisions and instances of being trapped. }
\begin{tabular}{ccccc}
\hline
Inputs   & Methods        & 
 \begin{tabular}[c]{@{}c@{}}Avg. \\ Coll.\end{tabular} & \begin{tabular}[c]{@{}c@{}}Avg. \\ Trapped\end{tabular} &
 \begin{tabular}[c]{@{}c@{}}Env. \\ Note\end{tabular}   \\ 
\hline
LiDAR   & $\pi_{L}$     & 0     & 0 & No Smoke, GT  \\
        & $\pi_{cl}$    & 0     & 0 & No Smoke, GT  \\
   & $\pi_{L}$   & 0     & 3 & Trapped in Smoke \\
\hline
mmWave    & $\pi_{cl}$         & -  & - & Unable to navigate \\
    & $cGAN+\pi_{cl}$    & 1.60  & 1.60 & Collision and trapped \\
  & $\pi_{L}$         & 1.20  & 0.60 & Baseline - w/o CM \\
        & $cGAN+\pi_{L}$  & 0  & 1.17 & Baseline - generative \\
        & \textbf{$\pi_{con}$}  & \textbf{0}  & \textbf{0} & \textbf{Ours - contrastive} \\
\hline

\end{tabular}
\label{table:nav_compare}
\vspace{-12pt}
\end{table}

\subsubsection{Generative reconstruction baselines}

For each method, the test was run for five laps, and we recorded the number of collisions and instances of being trapped. We evaluated the classic $\pi_{cl}$ and RL $\pi_{L}$ approaches, comparing them with $\pi_{cl}$ and $\pi_{L}$ enhanced by a reconstruction method (i.e., cGAN) using direct raw mmWave inputs. We found that the local map constructed using raw mmWave inputs lacked occupancy, resulting in a failure to navigate from the use of $\pi_{cl}$. Conversely, through the use of the cGAN to reconstruct the range data, the robot could navigate using the classic planning method, albeit with considerable collisions and instances of being trapped due to reconstruction errors. The control policy $\pi_L$ was more robust to noisy raw mmWave inputs without any tuning of the neural network, and $\pi_L$ managed to navigate through the maze even without reconstruction. When $cGAN + \pi_L$ was used, the robot collided less frequently but got trapped more frequently. 

\subsubsection{Cross-modal contrastive learning of representations}

$\pi_{con}$ performed the best among all of the evaluated methods, where the deep RL control policy was used after training with contrastive loss. $\pi_{con}$, as an end-to-end network, allowed the robot to safely traverse the entire maze in all five trials without colliding or being trapped.



\vspace{-4pt}
\subsection{Navigation in a General Indoor Corridor}

To further evaluate the navigation performance in a general indoor environment, we conducted the second experiment in a corridor located at another building which was not used for training data collection. The environment has two major corridors of 30 and 50 meters, respectively, two $90^{\circ}$ turns and two narrow gates. For safety considerations, we did not emit smoke in this environment. The details of this environment are shown in Fig. \ref{fig:exp-maze}.

\begin{figure}[t]
    \includegraphics[width=0.48\columnwidth]{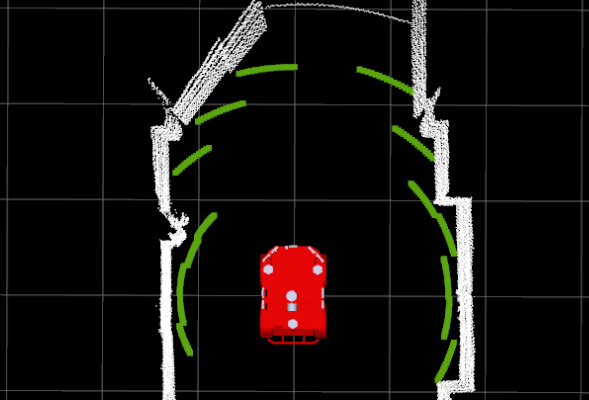}
    \includegraphics[width=0.48\columnwidth]{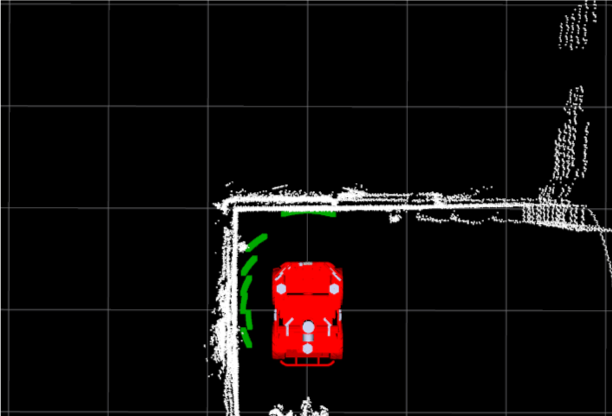}
    \caption{We used truncated LiDAR baseline to emulate sonar ring inputs (green line). The emulated sonar ring has $15^\circ$ resolution and 3 meters of maximum range. In the experiments we found that low angular resolution made it difficult to find an opening to navigate through narrow passage (left); short range is also a disadvantage of making turns (right).}
    \label{fig:sonar}
\end{figure}

\subsubsection{Truncated LiDAR (Emulated Sonar Ring) baseline }

In addition to previously mentioned baselines for navigation, we introduce a truncated LiDAR baseline to emulate a sonar ring to be compared against. In the review~\cite{Kleeman2008} a sonar ring was in the resolution of $15^{\circ}$ resolution and a maximum range of 3 meters. Following such settings, we carried out an experiment with decimated and truncated LiDAR measurements of 24 beams up to 3 meters. For the size of the Husky robot (67cm in width) low angular resolution  makes it difficult for the robot to find an opening in a relatively narrow passage (1.45 meters in width in our experiment). Short range measurement is also a disadvantage while making turns, shown in Fig.~\ref{fig:sonar}. The quantitative results in Table~\ref{table:nav_compare_edb1} showed that the robot was trapped at both of the narrow passages. 

\subsubsection{Generative reconstruction baselines}
With the help of the reconstruction model, both $\pi_{cl}$ and $\pi_L$ are able to navigate through the corridor with some collisions and instances of being trapped, but they happened more frequently than in the maze environment. We observed that the reconstruction model sometime generates corridor geometry with errors in angle compared to the ground truth when the vehicle is $10^{\circ} \sim 45^{\circ}$ relative to the wall. The errors in angle caused minor reconstruction loss but misled the vehicle to finally collide into the wall. This implies our dataset should be more diverse with more data of the robot not parallel to the walls to improve the generative reconstruction performance.

\subsubsection{Cross-modal contrastive learning}
$\pi_{con}$ also performed the best in this environment. Only one instance of being trapped happened among all five trials when robot navigate to the narrow gate in the environment.


\vspace{-8pt}
\begin{table}[ht]
\centering
\caption{Navigation performance at indoor corridor.}
\begin{tabular}{ccccc}
\hline
Inputs   & Methods        & 
 \begin{tabular}[c]{@{}c@{}}Avg. \\ Coll.\end{tabular} & \begin{tabular}[c]{@{}c@{}}Avg. \\ Trapped\end{tabular} &
 \begin{tabular}[c]{@{}c@{}}Env. \\ Note\end{tabular}   \\ 
\hline
LiDAR   & $\pi_{L}$     & 0     & 0 & No Smoke, GT  \\
        & $\pi_{cl}$    & 0     & 0 & No Smoke, GT  \\
Truncated & $\pi_{cl}$    & 0     & 2 & Emulate Sonar  \\
\hline
mmWave  & $\pi_{cl}$         & -    & - & Unable to navigate \\
        & $cGAN+\pi_{cl}$    & 3.2  & 1 & Collision and trapped \\
        & $\pi_{L}$          & 7    & 0 & Baseline - w/o CM \\
        & $cGAN+\pi_{L}$     & 0.6  & 0.6 & Baseline - generative \\
        & \textbf{$\pi_{con}$}  & \textbf{0}  & \textbf{0.2} & \textbf{Ours - contrastive} \\
\hline

\end{tabular}
\label{table:nav_compare_edb1}
\vspace{-16pt}
\end{table}

\subsection{General Discussions} \label{sec:general_discuss}

We found that the results of Table II and III are consistent that the proposed contrastive learning obtained better performance than baseline methods. The results may indicate that learning the representation may be more robust across different environments. We conjecture that for $cGAN + \pi_{L}$ to achieve the same performance as $\pi_{con}$, it is better for the reconstructed mmWave data to be similar to detailed LiDAR data. Without such similarity, the feature extraction network in $\pi_{L}$ may generate misleading representations for the decision-making network. The results of generative baseline in different environments were found to decrease  the number of trapped cases, and a slightly increased number of collisions in dense wall environments. The results may indicate that the generative reconstruction in different environments may be sensitive to the training dataset, which is consistent with the results of Table I. Future work could collect more training data in diverse environments to further verify the influences of training data to learn reconstruction.

We had tested two different mmWave sensor configurations on the same robot (Husky), shown in Fig. 2. Configuration 1 was used to collect training data and conduct the experiments, and configuration 2 was designed for potential use of smaller robots. We found that the different mmWave sensor configurations did not affect navigation performance on the same Husky robot. However, We suggest that additional transfer learning efforts considering robot size, speed, and reward settings may be needed to apply the method to a different robot.

An alternative to contrastive learning is hallucination, which has been used in other cross-modal tasks~\cite{Hoffman-2016-cvpr-modality-hallucination, Saputra-2019-ral-deeptio}. \cite{Hoffman-2016-cvpr-modality-hallucination} training a modality hallucination architecture to learn a multimodal convolutional network for object detection. The hallucination was trained to take an RGB input image and mimic the depth mid-level activations. Cross-modal studies on high-bandwidth inputs like images and low-bandwidth inputs (LiDAR/mmWave) could be further studied.

\vspace{-4pt}
\section{Conclusions} \label{Conclusions}
Learning how to navigate is difficult under adverse environmental conditions, where sensing data are degraded. As a remedy, we leveraged mmWave radars and proposed a method that uses cross-modal contrastive learning of representations  with both simulated and real-world data. Our method allows for automated navigation in obscurant-filled environments using data from only single-chip mmWave radars. In a comprehensive evaluation, we incorporated state-of-the-art generative reconstruction with a control policy for navigation tasks. Our method performed best in the smoke-filled environment against its counterparts, with results that were comparable to those from groundtruth LiDAR approaches in a smoke-free environment. Future studies can apply our method to learning across modes and platforms.

\vspace{-6pt}
\section*{Acknowledgments}
\label{sec:Acknowledgments}
The research was supported by Taiwan's Ministry of Science and Technology (grants 109-2321-B-009-006, 109-2224-E-007-004, and 109-2221-E-009-074). This work was funded in part by Qualcomm through the Taiwan University Research Collaboration Project.

\bibliographystyle{IEEEtran}
\bibliography{references.bib}

\end{document}